# Guiding Generative Models to Uncover Diverse and Novel Crystals via Reinforcement Learning

Hyunsoo Park* and Aron Walsh*

Department of Materials, Imperial College London, London SW7 2AZ, UK

Email: h.park@imperial.ac.uk and a.walsh@imperial.ac.uk

## Abstract

Discovering functional crystalline materials entails navigating an immense combinatorial design space. While recent advances in generative artificial intelligence have enabled the sampling of chemically plausible compositions and structures, a fundamental challenge remains: the objective misalignment between likelihood-based sampling in generative modelling and targeted focus on underexplored regions where novel compounds reside. Here, we introduce a reinforcement learning framework that guides latent denoising diffusion models toward diverse and novel, yet thermodynamically viable crystalline compounds. Our approach integrates group relative policy optimisation with verifiable, multi-objective rewards that jointly balance creativity, stability, and diversity. Beyond de novo generation, we demonstrate enhanced property-guided design that preserves chemical validity, while targeting desired functional properties. This approach establishes a modular foundation for controllable AI-driven inverse design that addresses the novelty-validity trade-off across scientific discovery applications of generative models.

## Introduction

The discovery of crystalline materials has historically shaped technological revolutions, from semiconductors[1], battery materials[2], and catalysts for energy and environmental applications[3]. Yet, the design space of crystalline compounds is vast and combinatorially complex, with the overwhelming majority remaining unexplored.[4] The emergence of artificial intelligence (AI) marks a substantial shift in materials design, offering the ability to navigate the high-dimensional composition-structure-property landscape with unprecedented efficiency and systematic exploration that extends far beyond the reach of conventional methodologies.[5,6]

Recent progress in generative modelling has demonstrated the ability to sample chemically plausible crystalline compounds[7], spanning variational autoencoders (VAEs)[8,9], generative adversarial networks (GANs)[10,11], autoregressive Transformer models[12–14], and continuous-time approaches such as score-based (diffusion) models[5,15,16] and flow matching[17,18]. Despite these advances, a fundamental misalignment exists between materials discovery and generative modelling. Generative models are typically trained to maximise likelihood[19,20], learning to reproduce statistical patterns of the training data. Conversely, the goal of materials discovery is to navigate systematically toward regions where chemical novelty and thermodynamic viability converge. These are precisely the underexplored spaces, often sparsely represented in existing databases, where viable yet unprecedented compounds are most likely to reside. This objective mismatch constrains the capacity of likelihood-based approaches to systematically explore chemical spaces of greatest scientific interest.

Overcoming this limitation requires generative models that can explore chemical space guided by explicit design objectives rather than statistical similarity to known data. Reinforcement learning (RL) provides a principled framework for this goal by optimising expected rewards instead of the likelihood of the data. Post-training with RL algorithms such as proximal policy optimisation (PPO)[21] has enabled generative models to align their outputs with predefined objectives and exhibit controllable behaviours across diverse domains, most notably in language[22] and vision[23]. However, applications to crystalline compounds remain limited. The periodic three-dimensional geometry, the coupling between lattice parameters and fractional coordinates, and symmetry constraints complicate the definition of tractable action spaces, and stable credit assignment throughout a generation trajectory during RL training.

In this work, we introduce a policy gradient RL framework tailored for denoising diffusion models in the latent space of crystalline structures. To ensure stable optimisation, we adopt the group relative policy optimisation (GRPO)[24] algorithm that performs group-normalised, preference-based updates to reduce gradient variance and integrate naturally with batched candidate generation. Exploration beyond the empirical data distribution is encouraged through a multi-objective reward design that jointly promotes creativity, validity, and diversity, effectively addressing the trade-off between novelty and stability that constrains conventional generative models. By enabling property-guided control without relying on heuristic mechanisms such as classifier-free guidance[25], our RL framework provides a principled pathway toward more controllable generation. Furthermore, the framework's modular design allows seamless alignment with diverse reward signals, establishing a versatile foundation for AI-guided inverse design of functional crystalline materials.

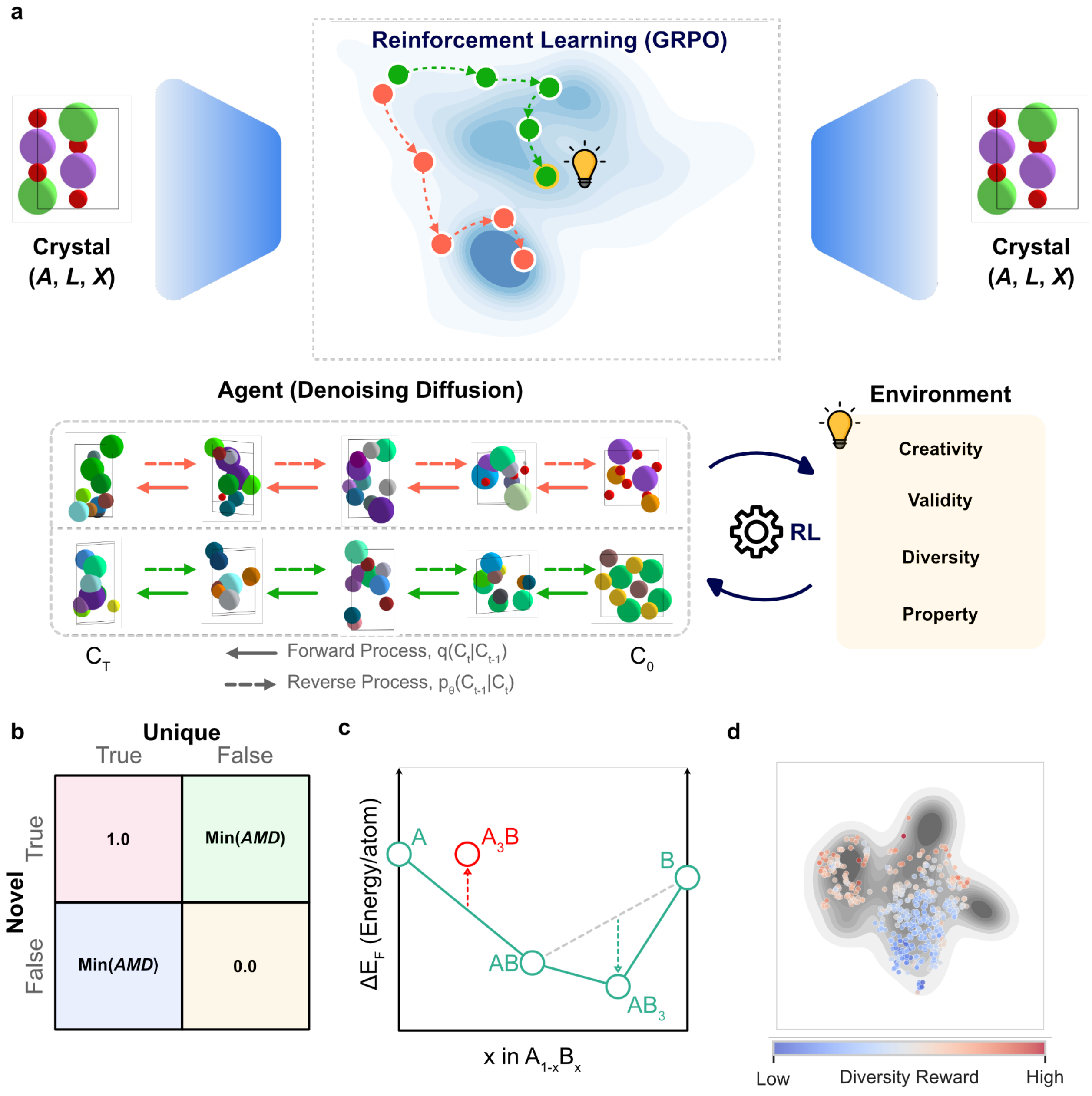

**Fig. 1. Reinforcement learning framework for guided exploration of crystalline materials.** (a) Illustration of the reinforcement learning framework in latent diffusion with group relative policy optimisation (GRPO) algorithm. (b) Continuous creativity reward assignment based on Average Minimum Distance (AMD), providing smooth gradients from 0 (neither unique nor novel) to 1 (both satisfied). (c) Stability reward mapping as a function of the predicted energy above the convex hull. (d) t-SNE projection of structural embeddings showing 1,000 reference (training) structures (grey density contours) and 500 generated structures coloured by diversity rewards.

# Results

## Reinforcement Learning Framework for Materials Space Exploration

Effective computational materials discovery strategies require not just producing plausible structures absent from databases, but in systematically accessing underexplored regions of materials space. Current generative AI models excel at interpolative novelty, creating plausible structures that statistically resemble training data. However, they face issues when extrapolating toward unfamiliar chemistries characterised by rare element combinations, unconventional coordination environments, or low-probability structural motifs. To address this challenge, we propose an RL framework that guides latent denoising diffusion models as illustrated in Fig. 1a.

Our framework integrates policy optimisation with generative models for crystalline compounds through two interconnected components. The *agent* comprises a latent denoising diffusion model serving as the policy network, while the *environment* encompasses reward evaluation mechanisms that assess generated structures across multiple objective functions. This architecture establishes a feedback loop whereby the generative model iteratively refines its exploration strategy based on explicit reward signals.

Prior to RL training, the *agent* undergoes pre-training on large-scale structure databases to learn essential crystallographic constraints, including local coordination patterns, periodicity, and symmetry-related geometries that characterise physically plausible compounds. During RL training, the agent updates parameters through on-policy gradient methods rather than score-matching objectives, transitioning from likelihood-based generation to reward-driven exploration.

## Latent Diffusion Model as Policy Network

Crystal structures are represented by atom types ($A$), lattice matrix ($L$), and fractional coordinates ($X$). Defining a tractable action space for direct manipulation of these components poses significant challenges to the application of reinforcement learning. Operating directly in the ($A, L, X$) space renders the policy's log probability intractable in policy gradient methods due to the joint updating of interdependent representations ($L$ and $X$), and the mixing of categorical variables ($A$) with continuous variables ($L$ and $X$).

To address these complexities, we adopt a latent diffusion model. Our approach is inspired by the All-atom Diffusion Transformer (ADiT)[26] architecture, where crystal structures are first encoded into a unified latent representation ($z$) using variational autoencoders (VAEs). VAEs create latent representations that facilitate capturing the coupled relationships among composition, geometry, and atomic arrangements. This encoding strategy transforms the heterogeneous $(A, L, X)$ space into a homogeneous continuous vector in a regularised latent space, rendering the action space tractable for policy gradient optimisation. Within this latent geometry, each denoising step corresponds to an action in which the policy network predicts a noise component that updates the latent state along a reverse diffusion trajectory. The decoder network subsequently reconstructs full crystal structures from these latent states.

## Reward Design to Guide Novel Compounds Discovery

Effective reward engineering is fundamental to successful reinforcement learning in materials discovery. The core challenge lies in designing reward signals that are simultaneously *verifiable*, grounded in physically meaningful quantities and *computational efficiency*, enabling rapid evaluation during iterative policy optimisation. To achieve a core objective to systematically explore materials space, we design a multi-objective reward function that balances three critical dimensions: creativity, stability, and diversity.

**Creativity Reward.** The creativity component evaluates whether sampled structures represent chemically distinct configurations through two complementary metrics: unique (uniqueness within the generated batch) and novel (absence from the reference training dataset). These metrics are usually defined as binary (True/False), with no gradient information, thereby producing a discontinuous reward landscape that hinders effective policy learning. To address this limitation, we introduce a continuous formulation based on the Average Minimum Distance (AMD)[27], a pairwise distribution metric that provides a smooth, differentiable measure of structural similarity suitable for gradient-based optimisation[28]. As illustrated in Fig. 1b, our reward assignment strategy provides smooth transitions where structures satisfying both criteria receive maximum reward (1.0), those meeting neither receive zero, and intermediate cases are scored according to their minimum AMD values, which are clipped to a maximum of 1.0.

**Stability Reward.** Thermodynamic feasibility constitutes an essential constraint in materials design, determining whether generated structures are chemically plausible and thermodynamically viable. While community-standard energy convex hull analysis provides stability assessment based on total energy analysis (Fig. 1c), the high computational cost of quantum mechanical calculations makes this approach computationally prohibitive for on-policy RL training. This limitation is overcome by leveraging pre-trained machine learning force fields (MLFFs) for efficient stability evaluation during training. Validation of the MLFF accuracy was performed by comparing energy above convex hull ($E_{hull}$) values for 1,000 sampled structures calculated via both DFT and MLFF prediction (Supplementary Fig. S1). The results demonstrate close agreement, with prediction errors consistently below 0.10 eV/atom within the metastable window, which is a threshold commonly adopted as a practical criterion for identifying synthetically accessible compounds.[29] This level of accuracy ensures reliable thermodynamic guidance while enabling the rapid iteration cycles essential for effective policy optimisation.

**Diversity Reward.** We define diversity as the extent to which generated structures collectively cover the support of a reference distribution, as defined in the corresponding structural and compositional embedding space (e.g., MP-20[30] or Alex-MP-20[5,31,32]). Conventional approaches typically assess diversity at the batch level using kernel-based distributional metrics such as Maximum Mean Discrepancy (MMD)[33]. While effective for quantifying global alignment, these methods lack the granularity to assign diversity scores to individual samples. To address this limitation, we introduce a marginal utility decomposition, which estimates each generated structure's incremental contribution to overall coverage by evaluating the change in kernel-based support upon its inclusion. This formulation enables more fine-grained policy guidance through per-sample diversity scores, each representing a structure's incremental individual contribution to the collective coverage of the reference distribution. This diversity reward plays a critical role in preventing mode collapse during RL training, where the policy might otherwise exploit local optima that satisfy creativity and stability but lack chemical diversity. Technical details on kernel selection, embedding representations, and reward computation are provided in the Methods section.

Fig. 1d illustrates the diversity reward landscape through a two-dimensional t-SNE projection of structural embedding features, comprising reference structures from the MP-20 dataset (visualised as grey density contours) and policy-generated structures (represented as coloured points). The contour intensity reflects local reference density, with darker regions corresponding to high-probability areas of the training distribution. Generated structures are coloured according to their assigned diversity rewards, ranging from blue (low-reward) to red (high-reward).

The spatial distribution of rewards reveals how our diversity scoring mechanism guides systematic exploration of chemical space by implicitly balancing two objectives: maximising coverage of the reference distribution and minimising redundancy among generated samples. High-reward structures (red) predominantly occupy regions adjacent to high-density areas of the reference distribution, as estimated from the kernel embedding, yet remain undersampled by other generated structures, thereby contributing to the expansion of the explored chemical landscape. Conversely, low-reward structures (blue) cluster in regions either saturated with generated samples or distant from the reference manifold, indicating redundancy or deviation from the target distribution, both of which are penalised in our formulation.

In practical implementation, the overall reward function integrates three weighted components (i.e. creativity, stability, and diversity), with diversity further decomposed into structural and compositional dimensions. This multi-objective formulation enables balanced optimisation across competing design criteria, facilitating systematic exploration of material space. The mathematical formulation and implementation details are provided in the Methods section.

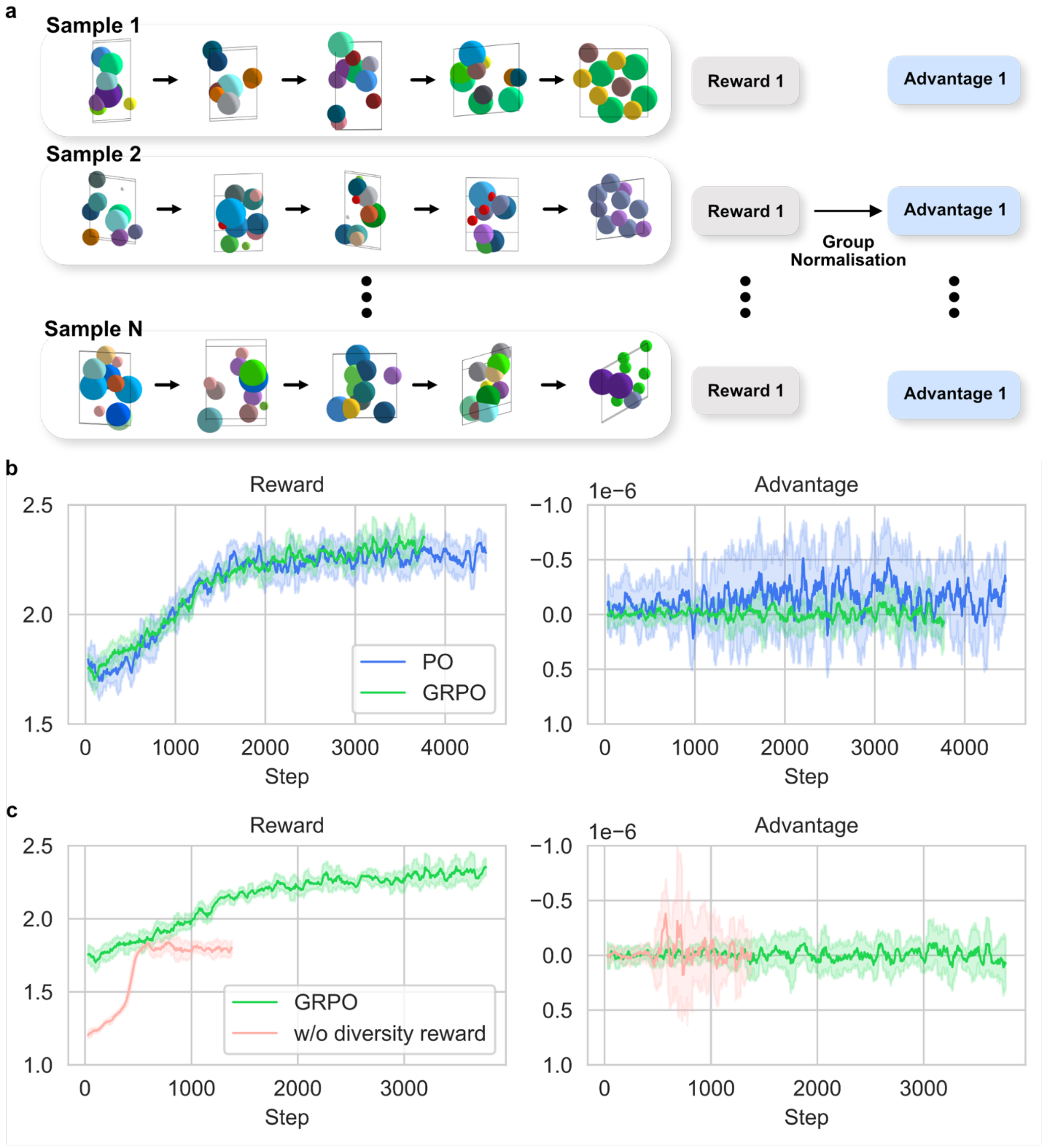

**Fig. 2. Illustration of the policy optimisation algorithm.** (a) Group Relative Policy Optimisation (GRPO) framework. (b) Ablation study of policy optimisation algorithm between GRPO and REINFORCE (PO) algorithm, where Step refers to the reinforcement learning step number. (C) Ablation study without diversity rewards in GRPO training.

## Policy Optimisation Algorithm

In policy gradient optimisation methods, we adopt GRPO to reduce gradient variance and enhance learning stability in batched candidate compounds generation. The fundamental policy-gradient algorithm is REINFORCE[34,35], which directly estimates policy gradients from sampled trajectories. However, REINFORCE suffers from high variance in gradient estimates, leading to unstable training dynamics and slow convergence. PPO was introduced to mitigate this issue by constraining policy updates within a clipped trust region, thereby improving training stability while retaining computational efficiency. Nevertheless, PPO typically requires a separate value network to estimate baseline advantages, introducing additional architectural complexity and training overhead.

GRPO addresses these limitations through a group-wise advantage normalisation strategy that reduces reliance on explicit value function estimation, as illustrated in Fig. 2a. In our framework, the generation process operates as follows: for each input condition (e.g., specified number of atoms in the unit cell), the latent diffusion model samples multiple candidate structures from the current policy. These structures form a group for which rewards are computed and subsequently normalised to obtain relative advantages. This group-based normalisation provides stable gradient signals by reducing sensitivity to absolute reward scales while maintaining the relative preferences that drive policy improvement. The technical details of the GRPO implementation are provided in the Methods section.

Fig. 2b compares the training dynamics of GRPO against the standard REINFORCE algorithm (hereafter, we refer to this as PO for simplicity). The PO, which computes advantages across mixed batches of candidate structures, exhibits substantially higher variance in advantage estimates during training. This higher variance correlates with structural variations in the batch, particularly in the number of atoms within each unit cell. It broadens the distribution of advantage values and conflates structural complexity with policy performance. In contrast, GRPO's group-wise normalisation confines advantage computation within subgroups of samples sharing the same number of atoms, yielding a markedly lower variance in gradient estimates. Both methods reach comparable rewards during training, with GRPO showing marginal improvements in the later stages of optimisation. These results suggest that group-wise normalisation primarily contributes to training stability, while its critical influence on overall performance is further reflected in the reward decomposition presented in Fig. 2c.

An ablation study was performed in which the diversity reward was removed while maintaining equal weights for the creativity and stability components. Without diversity reward, the policy exhibits severe mode collapse: generated structures converge to nearly identical chemical compositions, with subtle variations confined exclusively to lattice parameters and atomic coordinates (examples shown in Supplementary Fig. S2). This degeneracy arises because creativity rewards (uniqueness and novelty based on structural dissimilarity) and stability rewards (metastability) can be simultaneously maximised by repeatedly sampling minor geometric perturbations of a single favourable composition. The policy thus exploits this local optimum, producing structures that are technically unique and novel according to structural metrics, yet represent trivial variations lacking true chemical diversity. Incorporating the diversity reward alters this behaviour by discouraging compositional or structural redundancy, effectively preventing mode collapse during RL training.

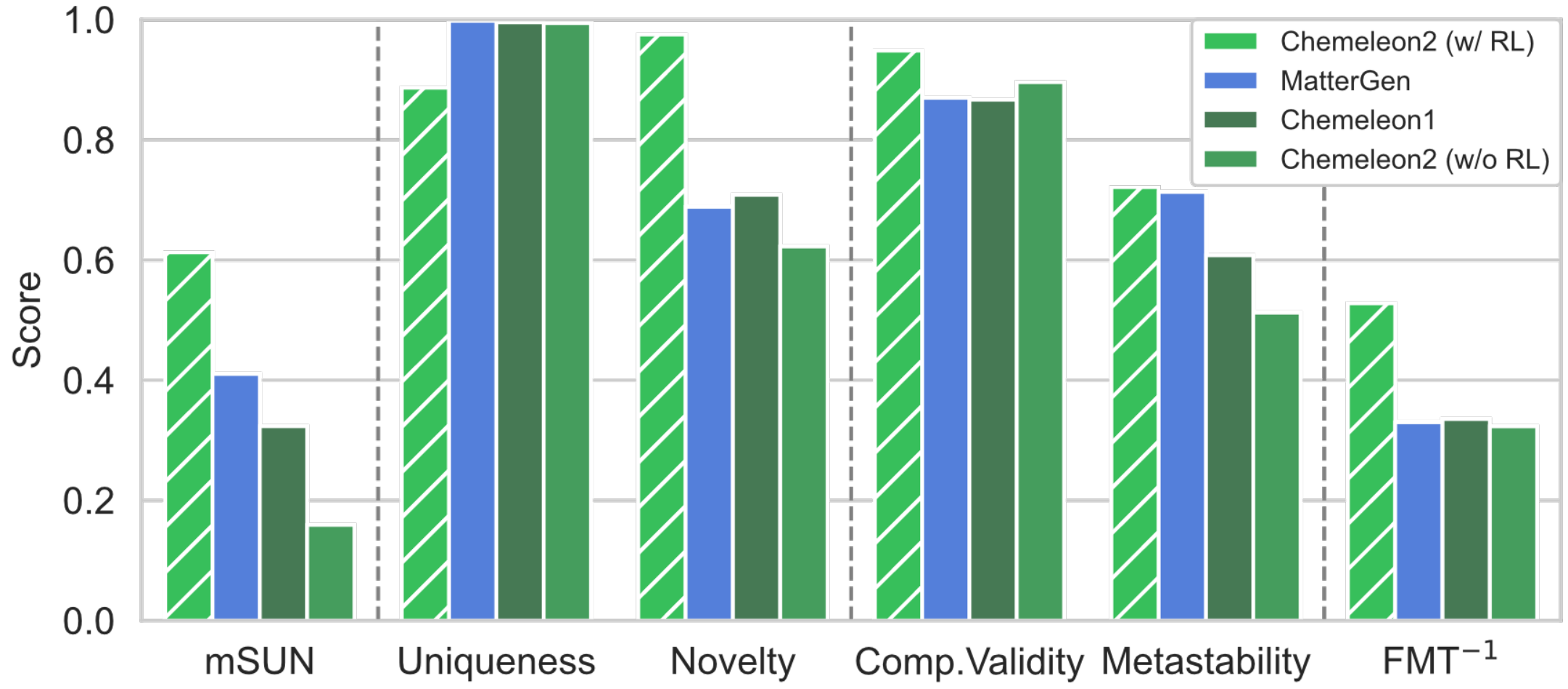

**Fig. 3. Comparison of generative models trained with the Alex-MP-20 dataset.** De novo crystal generation performance across models using 10,000 sampled structures. Metrics include mSUN (metastable, unique, and novel fraction), uniqueness, novelty, compositional validity (using SMACT filter), metastability ($\boldsymbol{E_{hull}^{MLFF}}$ < 0.1 eV/atom), and coverage (inverse Fréchet Materials Distance, FMD$^{-1}$).

## De Novo Generation Benchmark

Evaluation remains problematic due to the multifaceted nature of success criteria in generative modelling for materials design. We benchmark model performance in terms of creativity, validity, and coverage using a de novo generation (DNG) task comprising 10,000 crystal structures sampled per model. In this task, generation is conditioned solely on the number of atoms in the unit cell, and sampling is drawn from the distribution observed in the training dataset. Note that this setup differs from crystal structure prediction (CSP), which generates crystal structures from predefined chemical compositions.

To directly assess generative capability, all structures are evaluated as-generated without geometry optimisation. This approach captures whether models inherently learn to produce stable configurations while maintaining computational efficiency. Fig. 3 presents benchmark results for models trained on the large-scale Alex-MP-20 dataset, with analogous analysis on the standard MP-20 dataset provided in Supplementary Fig. S3 and Supplementary Table S1.

*Creativity* encompasses two dimensions as discussed in the reward design: the fraction of unique samples within the generated set (uniqueness), and the proportion of structures distinct from the reference dataset (novelty). *Validity* evaluates whether generated structures satisfy chemical plausibility in composition and thermodynamic feasibility in structure. Compositional validity employs SMACT[36,37] chemical filters that enforce charge neutrality and electronegativity compatibility through heuristic rules. Structural validity (or metastability) assesses thermodynamic plausibility via the fraction of structures exhibiting energy above the convex hull ($E_{hull}^{MLFF}$) below 0.1 eV/atom, which is a widely adopted threshold for synthetic accessibility. Stability evaluations leverage MLFFs for computational efficiency, with accuracy validated against DFT calculations in Supplementary Fig. S1. *Coverage* evaluates the inverse of the Fréchet Materials Distance ($FMD^{-1}$) computed on structural embeddings from the pre-trained VAE. This inversion reverses the metric's directionality, so that larger values correspond to closer distributional alignment between generated and reference structures, indicating improved coverage. Detailed formulations are provided in the Methods section.

Among these metrics, a widely adopted unified measure in generative modelling for materials is the mSUN score, which quantifies the fraction of generated structures that simultaneously satisfy metastable, unique, and novel. We use mSUN as the principal evaluation metric. Optimising against mSUN has the obstacle that increasing novelty and uniqueness typically drives generation toward unfamiliar chemical spaces where thermodynamic stability becomes uncertain, while prioritising metastability tends to confine sampling to well-explored regions of the training distribution. The further discussion of this trade-off is provided in the Discussion section.

Comprehensive analysis of models trained on the Alex-MP-20 dataset reveals clear performance variations across generative architectures (Figure 3). Among baseline models, MatterGen achieves the highest mSUN (41.0%), slightly exceeding its originally reported value of 38.6%, which utilised a workflow of DFT energy evaluations after geometry optimisation of each sampled structure. Chemeleon1[6], a text-guided diffusion model, was tailored for the DNG task by removing the text encoder, yielding an mSUN of 32.3 %. Both MatterGen and Chemeleon1, which operate directly on (graph-based) atomic representations rather than latent space, demonstrate exceptionally high uniqueness (exceeding 99.0 %) and strong metastability (71.3 % and 60.8 %, respectively) in their generated compounds.

The pre-trained Chemeleon2 baseline model, before RL training, shows substantially lower mSUN (15.9 %) with reduced novelty (62.3 %) and metastability (51.2 %) compared to the real-space diffusion models. This performance gap stems from architectural characteristics inherent to latent diffusion models. Chemeleon2 operates denoising in a compressed latent space via VAE, which produces smooth and regularised representations. This smoothness introduces two fundamental limitations. First, even minor perturbations or reconstruction errors in sparsely represented latent regions are amplified by the VAE decoder, often producing geometrically imprecise structures with distorted bond lengths and strained coordination environments that elevate formation energies and degrade metastability. Second, the regularised latent geometry systematically underrepresents rare or extreme structural configurations, biasing the diffusion process toward sampling familiar chemical spaces and thus constraining novelty. This behaviour aligns with the observed tendency of lower-capacity latent diffusion models (e.g., Chemeleon2 and ADiT) to exhibit reduced mSUN performance (see Supplementary Table S1).

Although latent diffusion models have some limitations, they offer key advantages in RL training: their continuous and homogeneous latent representations define tractable action spaces for policy-gradient optimisation. The integration of our RL framework, referred to as Chemeleon2 (w/ RL), achieves a marked enhancement in mSUN performance, increasing from 15.9 % to 61.3 %, representing an absolute gain of 45.4 % and a relative improvement of 285.5 % over the pre-trained baseline. Notably, RL training substantially elevated novelty from 70.8 % to 97.5 %, while also improving metastability from 51.2% to 72.1 %.

The decrease in uniqueness from 99.4 % to 88.7 % reflects an inherent challenge in on-policy reinforcement learning, which corresponds to a subtle mode collapse issue during training. Despite incorporating diversity rewards and entropy regularisation to mitigate this, the GRPO sampling strategy, which generates 64 structures per input condition during training, cannot fully prevent redundancy when scaling to 10,000 samples for evaluation.

Similar trends observed on the smaller standard MP-20 dataset (Supplementary Fig. S3) indicate that the RL framework elevates Chemeleon2's mSUN performance from 7.6 % to 26.0 %, achieving competitive performance to MatterGen (24.5 %). Notably, the RL-optimised Chemeleon2 achieves the highest novelty (96.5 %) among all benchmarked models, similar to the model trained with the larger Alex-MP-20 dataset. The more pronounced enhancement (from 15.9 % to 61.3 %) compared to MP-20 (from 7.6 % to 26.0 %) suggests that latent diffusion models benefit from RL-guided exploration when navigating larger configurational spaces.

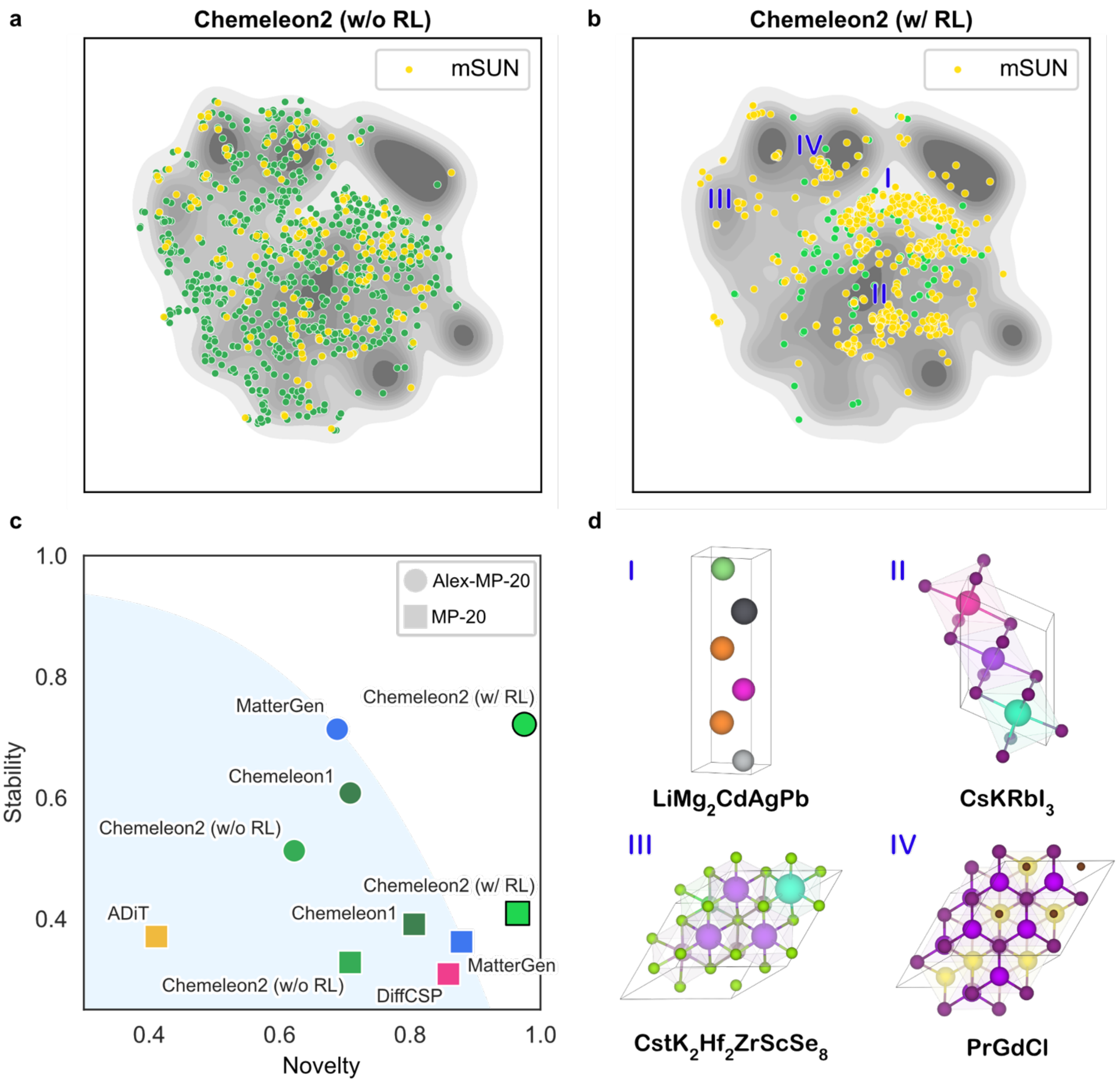

**Fig. 4. Novelty–stability dilemma in generative modelling for materials discovery.** t-SNE projections of structural embeddings for 1,000 generated samples from (a) the pre-trained Chemeleon2 baseline and (b) the Chemeleon2 with Reinforcement Learning (w/ RL) model. Grey contours denote the reference distribution from the Alex-MP-20 dataset, and yellow points indicate structures satisfying the mSUN criterion. (c) Trade-off between novelty and stability across generative models, highlighting the shift of the RL-optimised Chemeleon2 toward higher stability and novelty. (d) Representative crystal structures from four characteristic regions identified in (b).

# Discussion

## Novelty-Stability Dilemma

While generative modelling has emerged as a powerful tool to sample the space of unknown compounds, it is constrained by the novelty-stability dilemma. Compounds similar in chemistry to known materials (i.e. low novelty) are more likely to be stable compared to those that are dissimilar (i.e. high novelty). This is not a law of nature, but a common expectation in the materials chemistry domain. Indeed, many models for predicting the synthesisability of new compounds are based on distance measures with respect to the set of known materials.[38,39]

Fig. 4c illustrates this inherent trade-off across existing generative models, mapping novelty against stability (or metastability). Among baseline models, MatterGen lies on the Pareto frontier, achieving a balance between novelty in chemical composition or crystal structure and thermodynamic stability. Models such as MatterGen, DiffCSP, and Chemeleon1 employ diffusion-based frameworks tailored to different components of crystal structure representations. Typically, discrete denoising diffusion probabilistic model (D3PM)[40] handles categorical variables such as atomic species, while standard Denoising Diffusion Probabilistic Model (DDPM)[20] generates lattice matrices, and denoising score matching (DSM)[41] refines fractional coordinates, ensuring consistency with periodic boundary conditions for crystalline compounds.

Despite these architectural advances, a misalignment remains between conventional generative training objectives and the goals of materials design. Most generative models are trained to maximise data likelihood, thereby learning to sample from high-probability regions of the training distribution. In contrast, a materials discovery campaign aims to uncover chemically or configurationally novel structures that, by definition, occupy low-probability regions sparsely represented in existing databases. This tension gives rise to another definition of the novelty–stability dilemma for generative AI: unexplored chemical spaces (i.e. high novelty) yield synthetically inaccessible compounds, whereas prioritising stability confines the generation process to well-explored regions with high probability distributions similar to the training data (i.e. low novelty).

The RL strategy effectively mitigates this dilemma by encouraging exploration beyond the training distribution, while maintaining chemical accessibility. The Chemeleon2 (w/ RL) model extends the Pareto frontier, achieving both high novelty and high stability across small (MP-20) and large (Alex-MP-20) datasets. Such improvements cannot be achieved through conventional heuristic fine-tuning methods such as classifier-free guidance, which remain constrained by the underlying training data distribution and multi-objective trade-offs.

## Analysis of Generation Landscape

To visualise the distribution of training compounds and their relationship to those sampled by the models, Fig. 4a and 4b present t-SNE projections of structural embeddings for 1,000 samples. The background contours represent the density field of 10,000 structures randomly sampled from the reference Alex-MP-20 dataset, with darker regions indicating higher data concentration. Structures satisfying the mSUN criterion are highlighted in yellow. The t-SNE plots based on compositional and structural embeddings, coloured by energy distribution, are provided in Supplementary Fig. S4. Additional dimensionality reduction analyses, including PCA, t-SNE and UMAP on both compositional and structural embeddings, are presented in Supplementary Fig. S5–S7.

The Chemeleon2 baseline (Figure 4a) broadly covers the reference distribution, with generated structures dispersed across both high-density and peripheral regions. However, only a small fraction of these structures achieves mSUN (green and yellow points are scattered sparsely), indicating that while the model explores diverse regions according to their training distribution, it fails to systematically generate viable novel structures. Indeed, the $E_{hull}^{MLFF}$ distribution of baseline models exhibits a broad spread, with significant density extending beyond the metastability criterion of 0.1 eV/atom, as shown in Supplementary Fig. S8. This distribution reflects the tendency of likelihood-based models to sample proportionally to training data density without strategic preference. In contrast, the Chemeleon2 (w/ RL) shows selective spatial distribution, including lower density regions of the reference dataset, as shown in Figure 4b. The samples form compact clusters enriched in mSUN structures. This targeted sampling pattern is evidence of successful reward-driven optimisation, where the model has learned to preferentially explore regions of materials space that jointly maximise the targets.

To further illustrate how the RL policy guides systematic exploration in materials space, Fig. 4d, along with Supplementary Fig. S9, presents representative crystal structures from four primary regions.

**Region I** corresponds to the highest density of generated structures sampled from the RL-optimised model in Fig. 4b, where it predominantly contains multi-component intermetallic and complex metallic compounds (e.g., $Mg_2AgCdPb$, $SrCaCu_3ZnGe_4$, $PrGdAlGa_2Cu$, and $ZrSc_2CoRu_2$). These structures exhibit close-packed, high-coordination polyhedral arrangements, including distorted octahedral and cuboctahedral environments. While topologically similar to known prototype structures, these compounds target compositional novelty. They combine multiple elements with inequivalent crystallographic sites (e.g., two distinct Mg sites in $LiMg_2AgCdPb$; three Cu sites in $SrCaCu_3ZnGe_4$). We do note that such complex compositions could give rise to crystallographic disorder, where multiple species share a common sub-lattice. While this phenomenon goes beyond our present study, data-driven models are being developed to deal with such cases.[42,43]

**Region II** comprises soft-ionic, predominantly octahedral, frameworks built from alkali-centred $MI_6$ (M = Cs/Rb/K; X = I/Br) and rare-earth-centred $REX_6$ units that connect predominantly by corner and edge sharing with small octahedral tilts between 0 and 10°. The mixed-alkali halides (e.g., $CsRbKI_3$ and $CsKTlI_2Br$) display extended networks of corner/edge-sharing octahedra like $CsI_6$/$RbI_6$/$KI_6$. The mixed-anion rare-earth compounds (e.g., $TbHo_3ErTmSiTe_3As_2$ and $TbHoEr_3TmP_2CSe_3$) form dense $LnX_6$ nets (X = As/Te/Si or C/P/Se). The size-compatible cations and polarizable halide/chalcogenide anions stabilise slightly tilted octahedral networks with low-energy motifs consistently rewarded by the metastability criterion. Multi-cation (Cs/Rb/K) and multi-anion (I/Br, As/Te, C/P/Se) substitutions generate compositionally unique structures despite familiar local environments and connectivity patterns.

**Region III** comprises alkali-metal selenides ($Cs_2CeHfZrSe_6$, $CsK_3Hf_2ZrScSe_8$) and mixed-lanthanide chalcogenides ($Ce_2NdYErTm_2Se_8$, $HoErTm_4Sc_2C_3Se_4S$). These structures are constructed from well-defined $MSe_6$ octahedra (M = Ce/Hf/Zr/Sc/Tm/Er/Nd/Y) interconnected largely through edge-sharing, with alkali ions occupying 6-coordinate cavities.

**Region IV** is formed of lanthanide pnictide/halide/carbide compounds, including NdGdAsI, CeGdAsS, NdGdClI, and PrGdClI. They comprise nearly $LnX_6$ octahedra (X = As, S, I, Cl) with minimal tilting. Mixed-lanthanide occupancy (Nd/Gd, Ce/Gd, Pr/Gd) combined with mixed-anion coordination (As/S, Cl/I) produces distinctive first-shell chemistries and octahedral linkage schemes not captured by simple binary prototypes, despite the overall rocksalt-like topology.

The regional patterns reveal a systematic exploration strategy that the RL model learns and exploits. The policy network first identifies thermodynamically favourable structural prototypes and local motifs, then systematically introduces compositional complexity via controlled multi-element and multi-anion substitution patterns. This approach balances the competing objectives of stability and novelty, guiding towards solutions that may not be obvious based on knowledge of known materials alone.

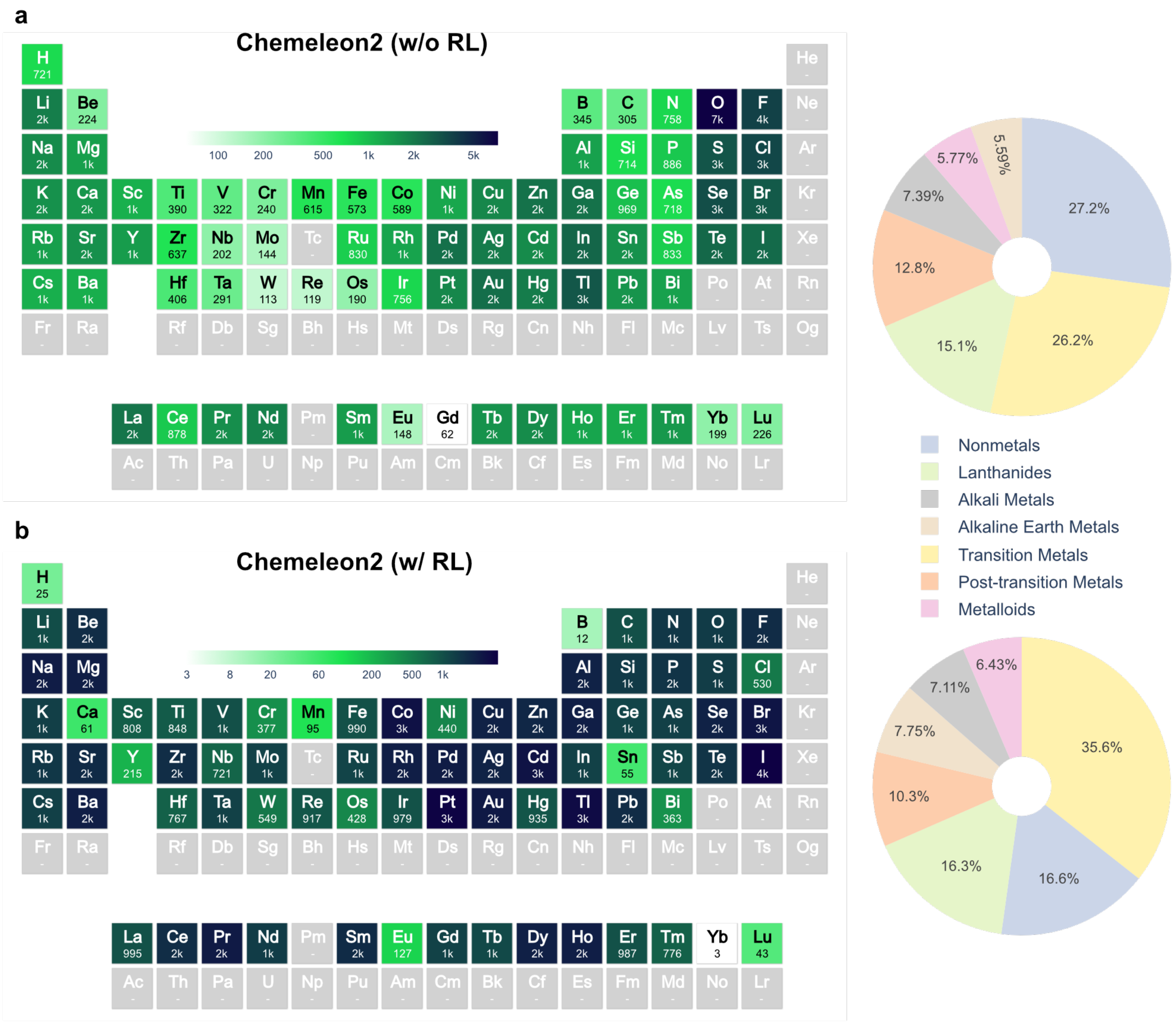

**Fig. 5. Statistical analysis of reinforcement learning–driven exploration of compositional space.** Elemental distribution heatmaps in periodic tables and corresponding pie charts showing the composition of 10,000 structures generated by (a) the pre-trained Chemeleon2 baseline and (b) the Chemeleon2 with reinforcement learning (w/ RL) model.

## Statistical Analysis of RL-Explored Compounds

It is well established that stable compounds are drawn from particular parts of the Periodic Table. For example, while most metals react with oxygen to form stable metal oxides, few stable crystals are formed of Noble gases under ambient conditions. Fig. 5 presents the elemental distribution statistics that capture compositional transformations induced by reinforcement learning across 10,000 sampled structures used for benchmarking. The pie charts indicate that the most substantial changes occur in the relative proportions of transition metals and non-metals. Transition metal content increases substantially from 26.2 % to 35.6 %, while non-metal content decreases correspondingly from 27.2 % to 16.3 %. More modest shifts are observed in other elemental categories: alkali earth metals, metalloids, and lanthanides exhibit slight increases, whereas alkali metals and post-transition metals show minor decreases, though these changes are less pronounced.

The baseline Chemeleon2 model preferentially generates compounds containing chalcogen and halogen, with a marked emphasis on oxygen (Supplementary Fig. S10). In contrast, the RL model yields a broader and more evenly distributed set of transition-metal chemistries, encompassing more polarizable species (I, Br, Se). This redistribution is consistent with the design of the multi-objective reward functions: the stability term promotes d-block elements with flexible oxidation states and ligand-field tunability, while the creativity and diversity rewards encourage mixing across multiple cation families and underrepresented elemental combinations. Analogous analyses for the MP-20 dataset are provided in Supplementary Fig. S11-12. The periodic tables and pie charts were generated using pymatviz[44].

## Flexible Reward Engineering and Property Guiding

Our RL framework features tunable control over the exploration–exploitation balance through adjustment of reward-function weights. Supplementary Fig. S13 shows the benchmark result of Chemeleon2 fine-tuned via RL training on the MP-20 dataset under two distinct weights ($w$ = 0.1, 1.0) for the diversity reward. Both configurations achieve comparable mSUN scores, yet the higher-weight model increases the $FMT^{-1}$ metric from 40.1 % to 71.2 %, accompanied by a moderate reduction in metastability from 40.9 % to 32.2 %. As illustrated in Supplementary Fig.

S14, t-SNE projection of structural embeddings qualitatively reveals a more uniformly dispersed distribution when diversity weighting is intensified. These findings suggest that, beyond the specific reward components used in this study, the modular reward design enables straightforward incorporation of additional verifiable objectives.

While de novo structure generation establishes the foundation for generative materials design, practical applications demand controllable synthesis of compounds tailored for target functional properties. To assess our framework's adaptability for property-driven design, we explore bandgap-targeted generation as a representative inverse design task. We benchmark the RL approach against classifier-free guidance (CFG)[25], which interpolates between conditioned and unconditioned modes through a tunable guidance scale (see Fig. 6a). Given that CFG often exhibits limitations in extrapolative regimes, producing off-manifold structures that violate physical constraints, the parameter-efficient adaptation via Low-Rank Adaptation (LoRA)[45] was applied to the diffusion transformer architecture during CFG fine-tuning to preserve the inductive biases learned during pre-training. In this setup, the pre-trained backbone remains frozen while low-rank adapter modules enable targeted property optimisation, avoiding catastrophic forgetting of pre-training knowledge.

For the benchmark of conditioned generation, 43,295 bandgap values provided in Alex-MP-20 dataset were utilised. The target bandgap was set to 3 eV, representing a deliberately challenging out-of-distribution objective, as shown by the bandgap distribution in Supplementary Fig. S15. Consistent with our generation methodology, all evaluations were conducted on as-generated structures without geometry optimisation, ensuring that the observed property distributions reflect intrinsic model capabilities rather than post-hoc structural refinement. A total of 512 structures were sampled from each model for comparative analysis. Further technical details on the property-guided generation process are provided in Supplementary Note S1.

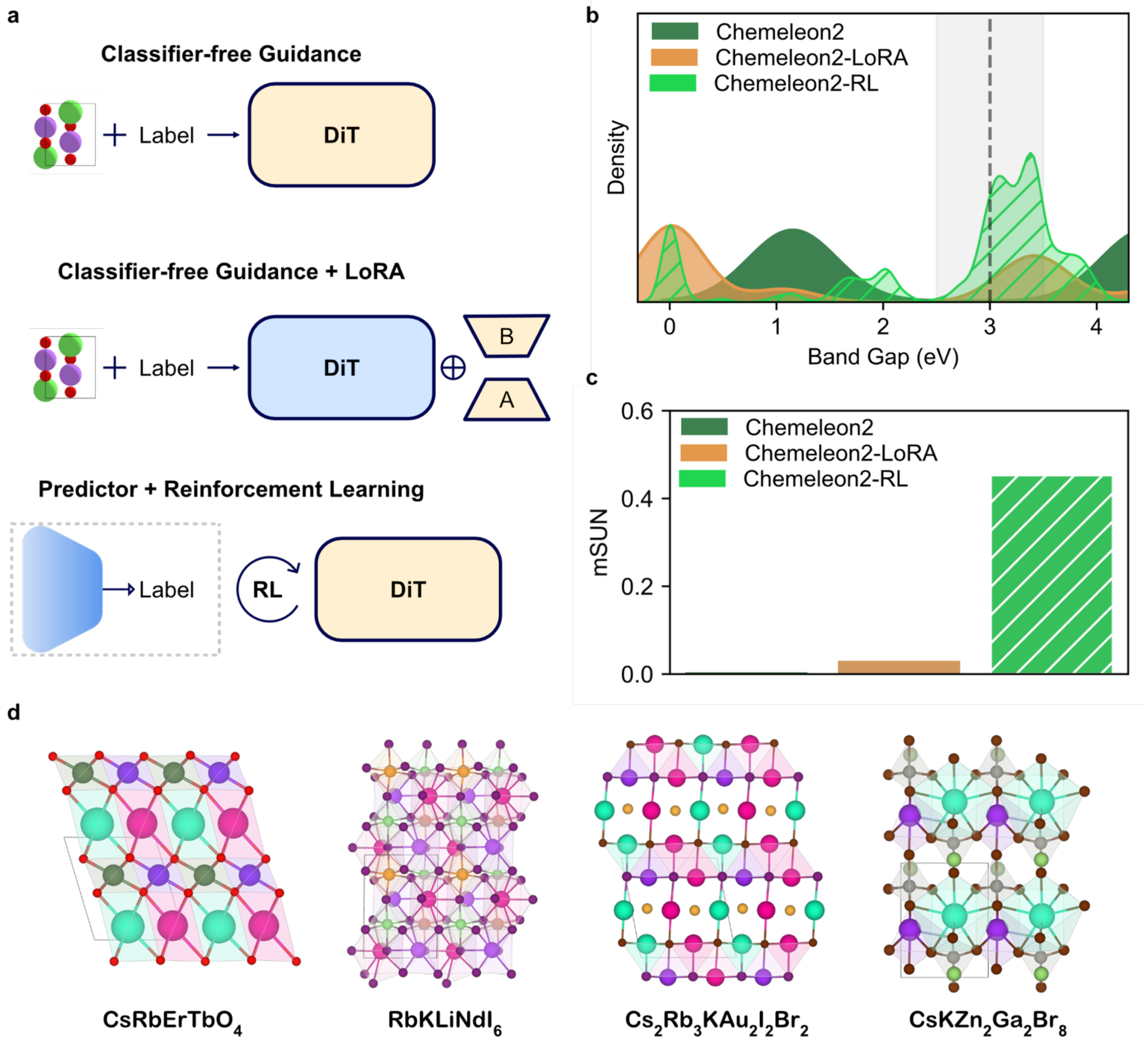

**Fig. 6. Reinforcement learning for property-guided generation.** (a) Schematic of property-conditioning strategies including classifier-free guidance (CFG), CFG with Low-Rank Adaptation (LoRA), and the reinforcement learning framework. (b) Bandgap distributions and (c) mSUN scores of 512 generated structures across different approaches, with the target bandgap set to 3 eV. (d) Representative four generated structures with target bandgaps near 3 eV.

For RL-based optimisation, we first constructed surrogate models to predict bandgap values from structural features extracted by the pre-trained VAE encoder. The surrogate, comprising simple fully connected layers, achieved mean absolute errors (MAEs) of 0.11 eV on the training set and 0.26 eV on the validation set, reflecting reliable predictive accuracy as a surrogate model for property optimisation (see Supplementary Fig. S16). Fig. 6b presents the distribution of DFT-calculated bandgaps across 512 generated structures, revealing marked differences in targeting precision. The CFG-guided models exhibit broad distributions spanning 0–5 eV, exhibiting weak clustering near the 3 eV target. In contrast, the RL-optimised Chemeleon2 exhibits a pronounced peak concentration around 3 eV, demonstrating enhanced property-guiding generation.

A notable feature across all methods is the prevalence of structures with computed bandgaps near 0 eV. It arises from the tough evaluation protocol, where all structures are assessed in their as-generated state without geometry relaxation. Unrelaxed configurations can contain strained bond lengths and distorted coordination environments that can induce spurious band overlap, thereby collapsing the computed bandgap to zero. When the analysis is restricted to structures satisfying the mSUN criterion, as shown in Supplementary Fig. S17, this zero-gap artefact is markedly diminished.

Maintaining the generation of chemically plausible and thermodynamically viable structures during property-guided fine-tuning represents a critical challenge in conditional materials design. Fig. 6c shows this trade-off by reporting the mSUN scores across different fine-tuning strategies. Chemeleon2 with CFG exhibits severe collapse of mSUN to near-zero values. The LoRA-adapted Chemeleon2 with CFG shows slight improvement, reaching only less 3%, which still reflects severe degradation. In contrast, MatterGen fine-tuning with CFG maintains mSUN at 27.6 %, representing modest degradation from its baseline performance as Supplementary Fig. S17c. These results indicate the instability of CFG-based finetuning when applied to compressed latent representations.

Note that the RL-optimised Chemeleon2 maintains mSUN at 45.3 %, demonstrating robust preservation of chemical validity alongside enhanced property controllability. Among the initial 512 generated samples, 82 mSUN structures having bandgaps within the target range of 2.7–3.3 eV. Four of the targeted compounds are illustrated in Fig. 6d. These are compositionally complex heteropolar crystals that obey charge balancing, e.g. the three monovalent (K, Rb, Li) cations and sole trivalent (Nd) cation balance the charge of the six iodides in $RbKLiNdI_6$. $CsKZn_2Ga_2Br_8$ adopts a complex structure based on Ga(I) to achieve charge balancing. Even $Cs_2KRb_3Au_2I_2Br_2$, which seems chemically unruly, can be broken down to a set of cations (Cs, K, Rb) and anions (Au, I, Br) distributed on a rocksalt crystal framework.

## Conclusions

This work establishes a framework for guiding generative models toward the systematic sampling of diverse and novel materials through reinforcement learning. By combining latent denoising diffusion models with GRPO-based optimisation and incorporating scientific objectives into verifiable multi-objective rewards, we achieve high performance in the mSUN metric with controllable balance between novelty, stability, and diversity. This integration in the open-source Chemeleon2 code enables the generation of thermodynamically viable crystalline compounds that systematically explore underrepresented regions of materials space. Critically, our framework addresses the novelty-stability dilemma that constrains conventional likelihood-based generative models, expanding the Pareto frontier through elaborate reward-driven exploration rather than pure distributional learning. Beyond de novo generation, we demonstrate property-guided design capabilities of latent diffusion that preserve chemical validity while precisely targeting functional characteristics.

The fundamental challenge we address in guiding generative models towards novel yet viable candidates rather than statistical reproduction of training data, extends beyond crystalline materials. Drug discovery, protein design, and molecular generation all face similar objective misalignment between likelihood-based training and exploration goals. Our modular reward framework provides a general strategy: explicit, verifiable objectives optimised through policy gradients can navigate the trade-offs inherent to scientific discovery applications of generative AI.

# Method

## Variational Autoencoder

The Chemeleon2 framework employs VAE as its core architecture for learning continuous latent representations of crystal structures. A crystal structure is formally defined as $C = (A, L, X)$, where $A = (A_1, A_2, \dots, A_N) \in \mathbb{Z}^N$ denotes the atomic species of $N$ atoms, $X = [x_1, x_2, \dots, x_n]^T \in [0,1)^3$ denotes the fractional coordinates of the atoms within the unit cell, and $L \in \mathbb{R}^{3\times3}$ is lattice matrix defining the unit cell dimensions and angles.

The VAE models the distribution $p(C)$ through a latent variable $z$ drawn from a prior distribution $p(\boldsymbol{z})$, typically assumed to follow a standard Gaussian distribution $\mathcal{N}(0, I)$. The encoder network, parameterised by $\phi$, approximates the posterior distribution $q_\phi(z \mid C)$. The decoder network, parameterised by $\theta$, defines $p_\theta(C \mid z)$ and performs the inverse transformation by reconstructing crystal structures from their latent encodings.

Training of the VAE proceeds by maximising the evidence lower bound (ELBO), which is given by:

$$\mathcal{L}(\theta, \phi) = \mathbb{E}_{q_\phi(z|C)}[\log\ p_\theta(C \mid z)] - \beta\ \mathrm{KL}[q_\phi(z \mid C) \parallel p(z)]$$

where the first term quantifies the reconstruction fidelity, and the second term enforces regularisation through the Kullback–Leibler (KL) divergence. The reconstruction component evaluates how accurately the decoder recovers the original crystal structure from its latent representation. The reconstruction loss involves categorical cross-entropy loss for atomic species classification, mean squared error (MSE) for lattice parameters and atomic fractional positions. The KL divergence term acts as a regulariser by constraining the learned posterior distribution $q_\phi(z \mid C)$ to remain close to the Gaussian prior $p(z)$.

Both the encoder and decoder employ Transformer-based architectures[46] that use multi-head self-attention mechanisms. Since Transformer models inherently struggle to capture crystal symmetries, stochastic data augmentation was applied during training with random translations of fractional coordinates and random rotations of lattice matrices and Cartesian coordinates.[26] The latent dimensionality was fixed at 8, with the loss component weights set to 1.0 for atomic species and lattice lengths, 10.0 for lattice angles and fractional coordinates, and $10^{-5}$ for the KL divergence term. Optimisation was performed using the AdamW algorithm with a learning rate of $10^{-4}$ and a batch size of 256. Training was conducted for a maximum of 5000 epochs, incorporating early stopping with a patience of 1,000 epochs based on the validation loss.

Reconstruction performance, evaluated using the StructureMatcher class from Pymatgen[47], shows that the VAE achieves a reconstruction accuracy of 99.4 % on the MP-20 test set, substantially surpassing the 85.5 % Transformer baseline in ADiT. This improvement is attributed to the use of lattice matrix in the representations for crystal structures, whereas ADiT employs Cartesian coordinates instead of a lattice matrix as VAE inputs.

## Denoising Diffusion

Latent diffusion models (LDMs)[48] operate by learning to denoise data representations within a compressed latent space generated by the VAE. The denoising network in this framework adopts the Diffusion Transformer[49] architecture, implemented in ADiT, but differs in its generative mechanism, utilising a DDPM rather than a flow-matching approach.

In the forward diffusion process, the model gradually corrupts the latent variable $z_0$ over $T$ discrete time steps by sequentially adding Gaussian noise according to a predefined variance schedule $\{\beta_t\}_{t=1}^{T}$. This process is formally described as:

$$q(z_t \mid z_{t-1}) = \mathcal{N}(\sqrt{1-\beta_t}\, z_{t-1}, \beta_t I)$$

The reverse diffusion process aims to reconstruct $z_0$ from this noisy representation by sequentially denoising from $z_T$ to $z_0$. This reverse process is parameterised as $p_\theta(z_{t-1} \mid z_t)$, where a denoiser learns to predict the added noise at each step. Training proceeds by minimising the simplified DDPM objective:

$$\mathcal{L}_{\text{DDPM}} = \mathbb{E}_{t,z_0,\epsilon}[\| \epsilon - \epsilon_\theta(z_t, t) \|^2]$$

Here, $\epsilon$ denotes the Gaussian noise added in the forward process, while $\epsilon_\theta(z_t, t)$ represents the model's noise prediction.

During inference, the Denoising Diffusion Implicit Model (DDIM)[50] is utilised to enhance sampling efficiency. In contrast to DDPMs, the DDIM employs a deterministic denoising formulation that leverages the same learned noise predictions to generate samples with significantly fewer steps, while maintaining comparable quality. In practice, DDIM allows for efficient generation using only 50 denoising sampling, while the timestep T is set to 1,000 using DDPM process in training process.

The Diffusion Transformer model consists of a depth of 12 layers, a hidden dimension of 768, and 12 attention heads. All hyperparameters, including optimizer settings, learning rate, and batch size, same as those used for the VAE. The model was trained for 5,000 epochs without early stopping, using an exponential moving average (EMA) with a decay rate of 0.99.

## GRPO algorithm

We formulate diffusion-based crystal generation as a Markov decision process (MDP) over reverse-diffusion steps $t \in \{T, T-1, \ldots, 1\}$. Let $\mathrm{z}_t$ denote the noisy latent state at reverse step $t$ mapped into the continuous latent space from the VAE. The action $\mathrm{a}_t$ corresponds to the noise prediction $\epsilon_\theta(\mathrm{z}_t, t)$ that updates the latent state according to the DDPM reverse process.

The policy is given by the denoising transition kernel:

$$\pi_\theta(a_t \mid s_t) \equiv p_\theta(z_{t-1} \mid z_t, x)$$

where $x$ denotes optional conditioning constraints, including the number of atoms in the unit cell and target properties such as chemical composition, space groups. This formulation transforms the heterogeneous crystal representation space $(A, L, X)$ into a tractable, continuous action space. Unlike direct manipulation of the heterogeneous representation involving coupled discrete-continuous variables and intractable log-probability computations, the latent space provides a homogeneous representation where standard policy gradient methods can be applied efficiently.

The complete generation process involves two distinct stages. (1) The iterative denoising in latent space produces $\mathrm{z}_0$ from initial noise $\mathrm{z}_T$, followed by (2) the pre-trained VAE $D_\phi(c \mid z)$ decoding to reconstruct the full crystal structure $\mathrm{c}_0 = D_\phi(\mathrm{z}_0)$, where $\phi$ is frozen during RL training. Given that the multi-objective reward function $R$ is evaluated at $t = 0$ with a fully denoised sample $c_0$, the RL objective over trajectories $\tau = \{z_T, z_{T-1}, \dots, z_0\}$ in latent space is therefore:

$$J(\theta) = \mathbb{E}_{\tau\sim\pi_\theta}[R(c_0, x)] = \mathbb{E}_{\tau\sim\pi_\theta}\left[R(D_\phi(z_0), x)\right]$$

Using the baseline REINFORCE (likelihood-ratio) estimator, the policy gradient is given by:

$$\nabla_\theta J(\theta) = \mathbb{E}_{\tau\sim\pi_\theta}\left[\sum_{t=1}^{T} \nabla_\theta \log p_\theta\,(z_{t-1} \mid z_t, x)\; R(c_0, x)\right]$$

In diffusion-policy optimisation with terminal rewards, this score-function gradients can lead to unstable parameter updates due to sparse terminal rewards and the long temporal horizons characteristic of the diffusion process. PPO can help to address these issues by (i) clipping importance ratios to a trustable region and (ii) using a learned value function $V_\phi$ to construct advantages $A_t = G_t - V_\phi(s_t)$, where $G_t$ is a return estimator. Despite these established variance-reduction techniques, training a reliable critic (value function) over the high-dimensional diffusion states $c_t$ is computationally expensive and often misaligned with the underlying denoising dynamics.

GRPO replaces the critic with a group-relative baseline computed across multiple samples that share the same conditioning $x$. By contrasting rollouts that share the same conditioning $x$, GRPO exhibits a low-variance, on-policy baseline that sidesteps value-function misspecification, yielding more stable updates and better sample efficiency.

For each conditioning $x$, draw $G$ trajectories $\{\tau_i\}_{i=1}^{G}$ from the behaviour policy $\pi_{\theta_{\text{old}}}$, compute terminal rewards $r_i = R(D_\phi(z_0^{(i)}), x)$, and define the group-relative normalised reward $\hat{A}_i$:

$$\hat{A}_i = \frac{r_i - \text{mean}(r_{1:G})}{\text{std}(r_{1:G})}$$

Since rewards arrive only at termination, $\hat{A}_i$ is shared across all time steps of trajectory $i$. And the per-step importance ratio is computed in latent space as:

$$\rho_t(\theta) = \frac{p_\theta(z_{t-1} \mid z_t, x)}{p_{\theta_{\text{old}}}(z_{t-1} \mid z_t, x)}$$

GRPO then maximises a clipped surrogate as follows:

$$L_{GRPO}(\theta) \; = \; \mathbb{E}_x \, \mathbb{E}_{\{\tau_i\} \sim \pi_{\theta_{old}}} \left\{ \frac{1}{G} \sum_{i=1}^{G} \sum_{t=1}^{T} min\left[\rho_t(\theta) \, \hat{A}_i, \; clip(\rho_t(\theta), \, 1 - \epsilon, \, 1 + \epsilon) \, \hat{A}_i)\right] \right\}$$
$$- \, \beta D_{kl}\left[\pi_\theta \parallel \pi_{ref}\right] + \gamma \, H(\pi_\theta)$$

where $\varepsilon$ is the clipping parameter, $D_{\mathrm{KL}}$ is the KL divergence to a reference policy $\pi_{\mathrm{ref}}$, with unbiased estimator, $H$ is the policy entropy, and $\beta, \gamma$ are regularisation weights.

The first term represents the clipped surrogate objective, which computes the expected advantage-weighted probability ratio across all reverse diffusion steps, with the minimum operation enforcing a trust region that constrains the importance ratio $\rho_t(\theta)$ within $[1 - \varepsilon, 1 + \varepsilon]$ to prevent excessively large policy updates and ensure stable training dynamics. The second term is a KL divergence penalty between the current policy $\pi_\theta$ and a reference policy $\pi_{ref}$ (i.e., the pre-trained denoiser model). This regularisation term is critical for preserving the inductive biases acquired during pre-training, in particular the optimisation with diffusion models over trajectories, thereby mitigating rapid divergence of $\pi_\theta$ away from $\pi_{ref}$ towards chemically or physically implausible structures. Finally, the entropy bonus term is included to prevent mode collapse which encourages stochastic exploration by penalising deterministic policies, thereby maintaining sampling diversity.

In GRPO implementation, we sampled $G = 64$ rollouts with training proceeding using a batch size of 5 conditioning inputs $x$, yielding 320 total structures per gradient update. We used a conservative clipping parameter $\epsilon = 10^{-3}$, a strong KL regularisation weight $\beta = 1.0$, and a small entropy bonus $\gamma = 10^{-5}$. The policy parameters were optimized using AdamW with a learning rate of $10^{-5}$, employing gradient accumulation over 2 inner batches to stabilize updates. Training incorporated early stopping based on a 500-step reward plateau criterion. Overall, these conservative hyperparameter choices reflect the inherent challenges of policy optimisation over trajectories in high-dimensional continuous action spaces characteristic of latent diffusion models.

## Reward Engineering

To guide generative models toward systematic exploration of materials space, we formulate a multi-objective reward function that balances competing design criteria. Each reward component is normalised using min–max scaling so that all rewards lie within a comparable range between 0 and 1. The total reward is defined as a weighted linear combination of four components:

$$R_{total} = w_{creativity}R_{creativity} + w_{stability}R_{stability} + w_{comp.diversity}R_{comp.diversity} + w_{struct.diversity}R_{struc.diversity}$$

where $w_{creativity}, w_{stability}, w_{comp.diversity}, w_{struct.diversity}$ represent the respective weight for creativity, stability, compositional diversity, and structural diversity rewards. The weights were set to 1.0 for $w_{comp.diversity}$, 1.0 for $w_{stability}$, 1.0 for $w_{comp.diversity}$, and 0.1 for $w_{struct.diversity}$. This weighting scheme was determined through systematic hyperparameter optimisation to balance exploratory behaviour with thermodynamic viability. An ablation study investigating the effect of increasing $w_{struct.diversity}$ to 1.0 as provided in the Discussion section with Supplementary Figures S13–S14.

$\boldsymbol{R_{creativity}}$ The creativity reward assesses distinct sampled structures using two binary criteria: uniqueness within the generated batch and novelty relative to the reference dataset, as computed by *StructureMatcher* with its default parameter settings. Let $u_i, v_i \in \{0,1\}$ be the uniqueness and novelty indicators for a generated structure $x_i$. Since direct use of $(u_i, v_i)$ yields a discontinuous landscape, we employ AMD which is a continuous isometry-invariant descriptor for periodic crystals summarising local coordination under full periodic boundary conditions. It is constructed by averaging, over atoms in the (primitive) unit cell, the distances to their $j$-th nearest neighbours for $j = 1, \dots, k$. In our implementation, $k$ is set to 100.

Let $f(x)$ denote the reduced formula. For each generated structure $x_i$, comparisons are restricted to structures with the same reduced formula. Specifically, the comparison set includes (i) all other generated structures in the current batch and (ii) all structures in the reference dataset (e.g., MP-20, Alex-MP-20), excluding $x_i$ itself. The corresponding AMD vectors are compared using the Chebyshev distance, and the reward is defined as:

$$R_{creativity}(x_i) = \begin{cases} 1, (u_i = 1 \wedge v_i = 1) \\ 0, (u_i = 0 \wedge v_i = 0) \\ \min\limits_{y: f(y)=f(x_i), y \neq x_i} \Delta \| AMD(x_i) - AMD(y) \|_\infty, (otherwise) \end{cases}$$

This yields a smooth, differentiable signal for borderline cases where a structure is unique or novel but not both, while preserving clear binary outcomes for the obvious cases.

$\boldsymbol{R_{stability}}$ Thermodynamic stability is assessed via the energy above the convex hull ($E_{hull}$), which quantifies the driving force for decomposition into competing phases. The convex hull is constructed from reference formation energies in the Materials Project[30] database, computed using GGA/GGA+U[51] without correction to ensure compatibility with conventionally trained foundational MLFFs. The MACE-MPA-0[52] model was employed, with reliable accuracy validated against DFT calculations (Supplementary Fig. S1).

For a given generated structure $x_i$, the formation energy is computed as:

$$E_{form}^{MLFF}(x_i) = E_{total}^{MLFF}(x_i) - \sum_a \mu_a n_a(x_i)$$

where $n_a(x_i)$ counts atoms of element $a$ in $x_i$ and $\{\mu_a\}$ are elemental reference energies. The energy above hull is then determined by comparing $E_{form}^{MLFF}(x_i)$ to the convex hull of competing phases with the same composition.

And the stability reward is defined as:

$$R_{stability} = -clip(E_{hull}^{MLFF}, 0, 1)$$

where the clipping operation bounds the penalty between 0 and 1. Structures on or below the convex hull ($E_{\text{hull}} \leq 0$) receive zero penalty, while those with $E_{\text{hull}} \geq 1$ eV/atom receive the maximum penalty. This formulation provides smooth gradients within the clipping range.

$\boldsymbol{R_{diversity}}$ The diversity reward is computed through a kernel-based distributional alignment framework that operates on both composition and structure in a shared latent space obtained from a pre-trained VAE. The structural embeddings are the encoder's latent vectors of crystal structures, and the compositional embeddings are obtained by stoichiometric pooling over the atom-type embedding layer. At each policy step, a batch of generated structures in latent space is considered, $\mathrm{X}_g = \{z_g^1, \dots, z_g^M\}$ in latent space and compare against a pre-computed reference set $\mathrm{X}_r = \{z_r^1, \dots, z_r^N\}$, which represents training dataset (e.g., MP-20, Alex-MP-20). The reward is the negative, unbiased mixed MMD with a polynomial kernel $K(x, y) = (x^\top y + c)^d$:

$$r_{\text{div}}\left(\mathrm{X}_g; \mathrm{X}_r\right) = -\frac{1}{M(M-1)} \sum_{i \neq j}^{M} K\left(z_g^i, z_g^j\right) - \frac{1}{N(N-1)} \sum_{i \neq j}^{N} K\left(z_r^i, z_r^j\right) + \frac{2}{MN} \sum_{i=1}^{M} \sum_{i=1}^{N} K(z_g^i, z_r^i)$$

This objective is maximised during policy-gradient RL training. The first term penalises redundancy within the generated batch, the second normalises against the intrinsic diversity of the reference set, and the third term promotes alignment with the target distribution.

To provide dense supervision for policy-gradient updates, we use a marginal-utility reward that attributes credit to each sample by measuring its incremental contribution to the batch objective:

$$\hat{r}_{\text{div}}(z_g^m) = r_{\text{div}}(\mathrm{X}_g) - r_{\text{div}}(\mathrm{X}_g \setminus \{z_g^m\})$$

This leave-one-out formulation yields each sample's incremental contribution to overall diversity, yielding distinct reward signals for every trajectory in the batch.

## Metrics

To benchmark generative-model performance on the DNG task, 10,000 crystal structures are sampled per model. Sampling is conditioned only on the number of atoms per unit cell drawn from the distribution of the corresponding reference set (i.e., MP-20 or Alex-MP-20). The uniqueness is defined as the fraction of generated structures that are non-duplicative within the sampled batch, and the novelty is the fraction that do not match any structure in the reference dataset. Both quantities are evaluated using *StructureMatcher* with default tolerance parameters. Compositional validity is assessed with SMACT, using the ICSD24 oxidation-state set to verify charge neutrality and electronegativity compatibility. Structural validity (metastability) quantifies thermodynamic feasibility as the fraction of samples whose energy above the convex hull is below 0.1 eV per atom, following the same protocol as $R_{stability}$.

For distributional alignment, the Fréchet Materials Distance (FMD) is computed using structural embeddings obtained as latent vectors from a pre-trained VAE encoder of crystal structures. Let $X_g = \{z_g^1, \dots, z_g^M\}$ and $X_r = \{z_r^1, \dots, z_r^N\}$ denote embeddings for generated and reference sets, with means and covariances $\Sigma_g, \Sigma_r$. The FMD is defined as:

$$\mathrm{FMD}(X_g; X_r) = \| \mu_g - \mu_r \|_2^2 + \mathrm{Tr}(\Sigma_g + \Sigma_r - 2(\Sigma_g \Sigma_r)^{\frac{1}{2}})$$

This metric quantifies the Fréchet distance between two multivariate Gaussian distributions fitted to the embedded representations. To obtain a bounded coverage score where higher values indicate better alignment, the coverage score is calculated by:

$$\mathrm{FMD}^{-1} = \frac{1}{1 + \mathrm{FMD}(X_g; X_r)}$$

such that larger values correspond to closer alignment with the reference distribution in structural latent space.

## First-principles calculations

First-principles calculations were performed using the Vienna Ab initio Simulation Package (VASP)[53,54] with the Projector Augmented Wave[55] formalism to validate machine learning force field predictions (Supplementary Fig. S1) and to compute electronic bandgaps for property-guided generation benchmarks. All calculations employed the Perdew-Burke-Ernzerhof exchange-correlation functional[56]. To ensure consistency with the Materials Project database for energy convex hull analysis, computational workflows were constructed using Atomate2[57] and executed via Jobflow-remote[58], specifically utilising the MPGGAStaticMaker. This protocol ensures that formation energies and stability assessments remain directly comparable with the reference thermodynamic database used throughout this work, which we note will have limitations for magnetic and highly-correlated materials.

## Data and Code Availability

The source code for Chemeleon2 is publicly available at https://github.com/hspark1212/chemeleon2, including implementations of the reinforcement learning framework, latent diffusion model architecture, and reward engineering modules. The repository also contains all sampled crystal structures used in the benchmarking analyses presented in this work. Training metrics, hyperparameter configurations, and testing logs are accessible through Weights & Biases at https://wandb.ai/hspark1212/chemeleon2. The Chemeleon1 implementation is available at https://github.com/hspark1212/chemeleon-dng.

## Acknowledgements

We thank Masahiro Negeishi and Lars Schaaf for fruitful discussions and careful reading of the manuscript. This work was supported by EPSRC project EP/X037754/1. Via our membership of the UK's HEC Materials Chemistry Consortium, which is funded by EPSRC (EP/X035859/1), this work used the ARCHER2 UK National Supercomputing Service (http://www.archer2.ac.uk). We are grateful to the UK Materials and Molecular Modelling Hub for computational resources, which is partially funded by EPSRC (EP/T022213/1, EP/W032260/1 and EP/P020194/1). We also acknowledge computational resources and support provided by the Imperial College Research Computing Service (http://doi.org/10.14469/hpc/2232).

## Author Contributions

H.P. and A.W. conceptualised this project and wrote the manuscript. H.P. developed the code and performed all research and analysis.

## Competing Interests

The authors declare no competing interests.

# References

1. Mikhailova, M. P., Moiseev, K. D. & Yakovlev, Yu. P. Discovery of III–V Semiconductors: Physical Properties and Application. *Semiconductors* **53**, 273–290 (2019).
2. Mizushima, K., Jones, P. C., Wiseman, P. J. & Goodenough, J. B. LixCoO2 (0<x<-1): A new cathode material for batteries of high energy density. *Mater. Res. Bull.* **15**, 783–789 (1980).
3. Li, H., Eddaoudi, M., O'Keeffe, M. & Yaghi, O. M. Design and synthesis of an exceptionally stable and highly porous metal-organic framework. *Nature* **402**, 276–279 (1999).
4. Park, H., Onwuli, A., Butler, K. T. & Walsh, A. Mapping inorganic crystal chemical space. *Faraday Discuss.* **256**, 601–613 (2025).
5. Zeni, C. *et al.* A generative model for inorganic materials design. *Nature* **639**, 624–632 (2025).
6. Park, H., Onwuli, A. & Walsh, A. Exploration of crystal chemical space using text-guided generative artificial intelligence. *Nat. Commun.* **16**, 4379 (2025).
7. Park, H., Li, Z. & Walsh, A. Has generative artificial intelligence solved inverse materials design? *Matter* **7**, 2355–2367 (2024).
8. Noh, J. *et al.* Inverse Design of Solid-State Materials via a Continuous Representation. *Matter* **1**, 1370–1384 (2019).
9. Zhu, R., Nong, W., Yamazaki, S. & Hippalgaonkar, K. WyCryst: Wyckoff inorganic crystal generator framework. *Matter* **7**, 3469–3488 (2024).
10. Kim, S., Noh, J., Gu, G. H., Aspuru-Guzik, A. & Jung, Y. Generative Adversarial Networks for Crystal Structure Prediction. *ACS Cent. Sci.* **6**, 1412–1420 (2020).
11. Zhao, Y. *et al.* High-Throughput Discovery of Novel Cubic Crystal Materials Using Deep Generative Neural Networks. *Adv. Sci.* **8**, 2100566 (2021).
12. Antunes, L. M., Butler, K. T. & Grau-Crespo, R. Crystal structure generation with autoregressive large language modeling. *Nat. Commun.* **15**, 10570 (2024).
13. Gruver, N. *et al.* Fine-Tuned Language Models Generate Stable Inorganic Materials as Text. Preprint at https://doi.org/10.48550/arXiv.2402.04379 (2025).
14. Breuck, P.-P. D., Piracha, H. A., Rignanese, G.-M. & Marques, M. A. L. A generative material transformer using Wyckoff representation. Preprint at https://doi.org/10.48550/arXiv.2501.16051 (2025).
15. Xie, T., Fu, X., Ganea, O.-E., Barzilay, R. & Jaakkola, T. Crystal Diffusion Variational Autoencoder for Periodic Material Generation. Preprint at https://doi.org/10.48550/arXiv.2110.06197 (2022).
16. Jiao, R. *et al.* Crystal Structure Prediction by Joint Equivariant Diffusion. Preprint at https://doi.org/10.48550/arXiv.2309.04475 (2024).
17. Miller, B. K., Chen, R. T. Q., Sriram, A. & Wood, B. M. FlowMM: Generating Materials with Riemannian Flow Matching. Preprint at https://doi.org/10.48550/arXiv.2406.04713 (2024).
18. Luo, X. *et al.* CrystalFlow: a flow-based generative model for crystalline materials. *Nat. Commun.* **16**, 9267 (2025).

19. Kingma, D. P. & Welling, M. Auto-Encoding Variational Bayes. Preprint at https://doi.org/10.48550/arXiv.1312.6114 (2022).

20. Ho, J., Jain, A. & Abbeel, P. Denoising Diffusion Probabilistic Models. Preprint at https://doi.org/10.48550/arXiv.2006.11239 (2020).

21. Schulman, J., Wolski, F., Dhariwal, P., Radford, A. & Klimov, O. Proximal Policy Optimization Algorithms. Preprint at https://doi.org/10.48550/arXiv.1707.06347 (2017).

22. Lambert, N. Reinforcement Learning from Human Feedback. Preprint at https://doi.org/10.48550/arXiv.2504.12501 (2025).

23. Zhang, Y., Tzeng, E., Du, Y. & Kislyuk, D. Large-scale Reinforcement Learning for Diffusion Models. Preprint at https://doi.org/10.48550/arXiv.2401.12244 (2024).

24. Shao, Z. *et al.* DeepSeekMath: Pushing the Limits of Mathematical Reasoning in Open Language Models. Preprint at https://doi.org/10.48550/arXiv.2402.03300 (2024).

25. Ho, J. & Salimans, T. Classifier-Free Diffusion Guidance. Preprint at https://doi.org/10.48550/arXiv.2207.12598 (2022).

26. Joshi, C. K. *et al.* All-atom Diffusion Transformers: Unified generative modelling of molecules and materials. Preprint at https://doi.org/10.48550/arXiv.2503.03965 (2025).

27. Widdowson, D., Mosca, M. M., Pulido, A., Cooper, A. I. & Kurlin, V. Average minimum distances of periodic point sets – foundational invariants for mapping periodic crystals. *MATCH Commun. Math. Comput. Chem.* **87**, 529–559 (2022).

28. Negishi, M., Park, H., Mastej, K. O. & Walsh, A. Continuous Uniqueness and Novelty Metrics for Generative Modeling of Inorganic Crystals. Preprint at https://doi.org/10.48550/arXiv.2510.12405 (2025).

29. Aykol, M., Dwaraknath, S. S., Sun, W. & Persson, K. A. Thermodynamic limit for synthesis of metastable inorganic materials. *Sci. Adv.* **4**, eaaq0148 (2018).

30. Jain, A. *et al.* Commentary: The Materials Project: A materials genome approach to accelerating materials innovation. *APL Mater.* **1**, 011002 (2013).

31. Schmidt, J., Wang, H.-C., Cerqueira, T. F. T., Botti, S. & Marques, M. A. L. A dataset of 175k stable and metastable materials calculated with the PBEsol and SCAN functionals. *Sci. Data* **9**, 64 (2022).

32. Schmidt, J. *et al.* Large-scale machine-learning-assisted exploration of the whole materials space. Preprint at https://doi.org/10.48550/arXiv.2210.00579 (2022).

33. Gretton, A., Borgwardt, K., Rasch, M. J., Scholkopf, B. & Smola, A. J. A Kernel Method for the Two-Sample Problem. Preprint at https://doi.org/10.48550/arXiv.0805.2368 (2008).

34. Williams, R. J. Simple Statistical Gradient-Following Algorithms for Connectionist Reinforcement Learning. *Mach Learn* **8**, 229–256 (1992).

35. Mohamed, S., Rosca, M., Figurnov, M. & Mnih, A. Monte Carlo Gradient Estimation in Machine Learning. Preprint at https://doi.org/10.48550/arXiv.1906.10652 (2020).

36. Davies, D. W. *et al.* Computational Screening of All Stoichiometric Inorganic Materials. *Chem* **1**, 617–627 (2016).

37. Davies, D. W. *et al.* SMACT: Semiconducting Materials by Analogy and Chemical Theory. *J. Open Source Softw.* **4**, 1361 (2019).

38. Park, H., Mastej, K. O., Detrattanawichai, P., Nduma, R. & Walsh, A. Closing the synthesis gap in computational materials design. Preprint at https://doi.org/10.26434/chemrxiv-2025-sbc0c-v2 (2025).

39. Jang, J. *et al.* Synthesizability of materials stoichiometry using semi-supervised learning. *Matter* **7**, 2294–2312 (2024).

40. Austin, J., Johnson, D. D., Ho, J., Tarlow, D. & Berg, R. van den. Structured Denoising Diffusion Models in Discrete State-Spaces. Preprint at https://doi.org/10.48550/arXiv.2107.03006 (2023).

41. Song, Y. & Ermon, S. Improved Techniques for Training Score-Based Generative Models. Preprint at https://doi.org/10.48550/arXiv.2006.09011 (2020).

42. Petersen, M. H. *et al.* Dis-GEN: Disordered crystal structure generation. Preprint at https://doi.org/10.48550/arXiv.2507.18275 (2025).

43. Jakob, K. S., Walsh, A., Reuter, K. & Margraf, J. T. Learning Crystallographic Disorder: Bridging Prediction and Experiment in Materials Discovery. *Adv. Mater.* **n/a**, e14226.

44. Riebesell, J. *et al.* janosh/pymatviz: v0.17.2. Zenodo https://doi.org/10.5281/zenodo.16874341 (2025).

45. Hu, E. J. *et al.* LoRA: Low-Rank Adaptation of Large Language Models. Preprint at https://doi.org/10.48550/arXiv.2106.09685 (2021).

46. Vaswani, A. *et al.* Attention is All you Need. in *Advances in Neural Information Processing Systems* vol. 30 (Curran Associates, Inc., 2017).

47. Ong, S. P. *et al.* Python Materials Genomics (pymatgen): A robust, open-source python library for materials analysis. *Prof Ceder* https://dspace.mit.edu/handle/1721.1/101936 (2012).

48. Rombach, R., Blattmann, A., Lorenz, D., Esser, P. & Ommer, B. High-Resolution Image Synthesis with Latent Diffusion Models. Preprint at https://doi.org/10.48550/arXiv.2112.10752 (2022).

49. Peebles, W. & Xie, S. Scalable Diffusion Models with Transformers. in *2023 IEEE/CVF International Conference on Computer Vision (ICCV)* 4172–4182 (IEEE, Paris, France, 2023). doi:10.1109/ICCV51070.2023.00387.

50. Song, J., Meng, C. & Ermon, S. Denoising Diffusion Implicit Models. Preprint at https://doi.org/10.48550/arXiv.2010.02502 (2022).

51. Jain, A. *et al.* Formation enthalpies by mixing GGA and GGA \$+\$ \$U\$ calculations. *Phys. Rev. B* **84**, 045115 (2011).

52. Batatia, I. *et al.* A foundation model for atomistic materials chemistry. Preprint at https://doi.org/10.48550/arXiv.2401.00096 (2025).

53. Kresse, G. & Furthmüller, J. Efficient iterative schemes for ab initio total-energy calculations using a plane-wave basis set. *Phys. Rev. B* **54**, 11169–11186 (1996).

54. Kresse, G. & Furthmüller, J. Efficiency of ab-initio total energy calculations for metals and semiconductors using a plane-wave basis set. *Comput. Mater. Sci.* **6**, 15–50 (1996).

55. Blöchl, P. E. Projector augmented-wave method. *Phys. Rev. B* **50**, 17953–17979 (1994).

56. Perdew, J. P., Burke, K. & Ernzerhof, M. Generalized Gradient Approximation Made Simple. *Phys. Rev. Lett.* **77**, 3865–3868 (1996).

57. Ganose, A. M. *et al.* Atomate2: modular workflows for materials science. *Digit. Discov.* **4**, 1944–1973 (2025).

58. Rosen, A. S. *et al.* Jobflow: Computational Workflows Made Simple. *J. Open Source Softw.* **9**, 5995 (2024).

## Table of Contents

## Supplementary Note S1. Bandgap-guiding generation

This note provided detailed technical specifications as discussed in Property-guiding Generation section along with Figure 6. We compare three distinct conditioning strategies: classifier-free guidance (CFG), CFG with Low-Rank Adaptation (LoRA), and reinforcement learning (RL)-based optimisation.

### Classifier-Free Guidance (CFG)

The CFG baseline was implemented by fine-tuning the pre-trained latent diffusion model (LDM) on the Alex-MP-20 dataset with bandgap conditioning. The VAE encoder and decoder were frozen throughout training to preserve the learned latent geometry. The denoising diffusion transformer was trained to predict noise conditioned on both the number of atoms and target bandgap values.

Training proceeded for a maximum of 1,000 epochs using a batch size of 128 and a learning rate of $10^{-4}$. The AdamW optimizer was employed with initialisation from the pre-trained Alex-MP-20 denoiser model. The VAE encoder and decoder remained frozen throughout the fine-tuning process.

During inference, the guidance scale λ was set to 2.0, interpolating between conditional and unconditional predictions according to the standard CFG formulation:

$$\tilde{\epsilon}_\theta(z_t, t, c) = \epsilon_\theta(z_t, t, \emptyset) + \lambda(\epsilon_\theta(z_t, t, c) - \epsilon_\theta(z_t, t))$$

where $c$ represents the conditioning information (target bandgap). This interpolation mechanism enables control over the strength of property conditioning during the generation process.

### CFG with Low-Rank Adaptation (LoRA)

To mitigate catastrophic forgetting during property-guided fine-tuning, LoRA modules were incorporated into the diffusion transformer architecture. LoRA enables parameter-efficient adaptation by inserting low-rank decomposition matrices into selected layers while keeping the pre-trained backbone frozen. This approach preserves the inductive biases learned during pre-training while allowing targeted optimisation for property prediction.

The LoRA configuration employed a rank of 16 with a scaling factor of 32, which determines the dimensionality of the low-rank decomposition and the magnitude of the adapted weights, respectively. A dropout probability of 0.1 was applied to the LoRA layers to prevent overfitting.

LoRA modules were selectively applied to the attention output projection layer and both fully connected layers in the multi-layer perceptron blocks, specifically targeting the attention output projection and MLP fully connected layers.

**Bandgap Surrogate Model**

To train RL finetuning with bandgap condition, we need a surrogate model what provides rapid property estimation during policy optimisation. The surrogate architecture comprises a simple feedforward network with with SiLU (Sigmoid Linear Unit) activations and layer normalisation after each hidden layer. The predictor was trained with batch size 256, learning rate $10^{-3}$, and AdamW, minimising mean-squared error (MSE) between predicted and true bandgaps. We randomly split off 10% of the data as a validation set, yielding mean absolute errors (MAEs) of 0.11 eV for train and 0.26 eV for validation set.

**Reinforcement Learning Optimisation**

The RL framework for property-guided generation maintained the core GRPO algorithm described in the Methods section, with modifications to the reward function to incorporate bandgap targeting. Unlike the de novo generation task, which optimised for creativity, stability, and diversity, the property-guided variant focuses on precise functional targeting while maintaining compositional diversity.

The total reward is a weighted sum of a bandgap-targeting term and a compositional-diversity term:

$$R_{\text{total}} = w_{\text{bandgap}} \cdot R_{\text{bandgap}} + w_{\text{comp.div}} \cdot R_{\text{comp.diversity}}$$

The bandgap reward component employs a quadratic penalty function centred at the target value:

$$R_{\text{bandgap}}(x_i) = -(E_g^{\text{pred}}(x_i) - E_g^{\text{target}})^2$$

with $E_g^{\text{target}} = 3.0$ eV. We set $w_{\text{gap}} = 1.0$and $w_{\text{div}} = 0.5$, applying min–max normalisation to each reward term before weighting. The RL training implementation adhered to the same protocol and hyperparameters as the ones for DNG as provided in Methods section.

**Reproduce of MatterGen Results**

The MatterGen bandgap-conditioned model was evaluated using the publicly available checkpoint provided in their official GitHub repository. As shown in Supplementary Figure S17, the generated structures display a broad bandgap distribution spanning 0–5 eV, with limited density near the target bandgap of 3.0 eV. A substantial fraction exhibit computed gaps close to 0 eV. This observation requires careful interpretation in terms of the evaluation methodology.

In the original MatterGen study, the generation pipeline involves multiple post-processing stages: MLFF-based geometry optimisation, filtering by SUN criterion, DFT double relaxation, and final bandgap computation using BandStructureMaker. This comprehensive relaxation workflow effectively removes spurious zero-gap structures that can arise from non-equilibrium geometries.

In contrast, our benchmarking protocol used in this work, including Fig. 6 and Supplementary Fig. S17, applies single-point SCF calculations directly to the as-generated structures, without any geometric optimisation. Under this protocol, the RL-optimised Chemeleon2 model exhibits superior performance. Despite the absence of relaxation, RL-trained structures display a strong concentration near the target bandgap of 3.0 eV and a reduced zero-gap structures compared to CFG-based methods. A detailed discussion of reproduction challenges and methodological discrepancies is available at https://github.com/microsoft/mattergen/issues/213.

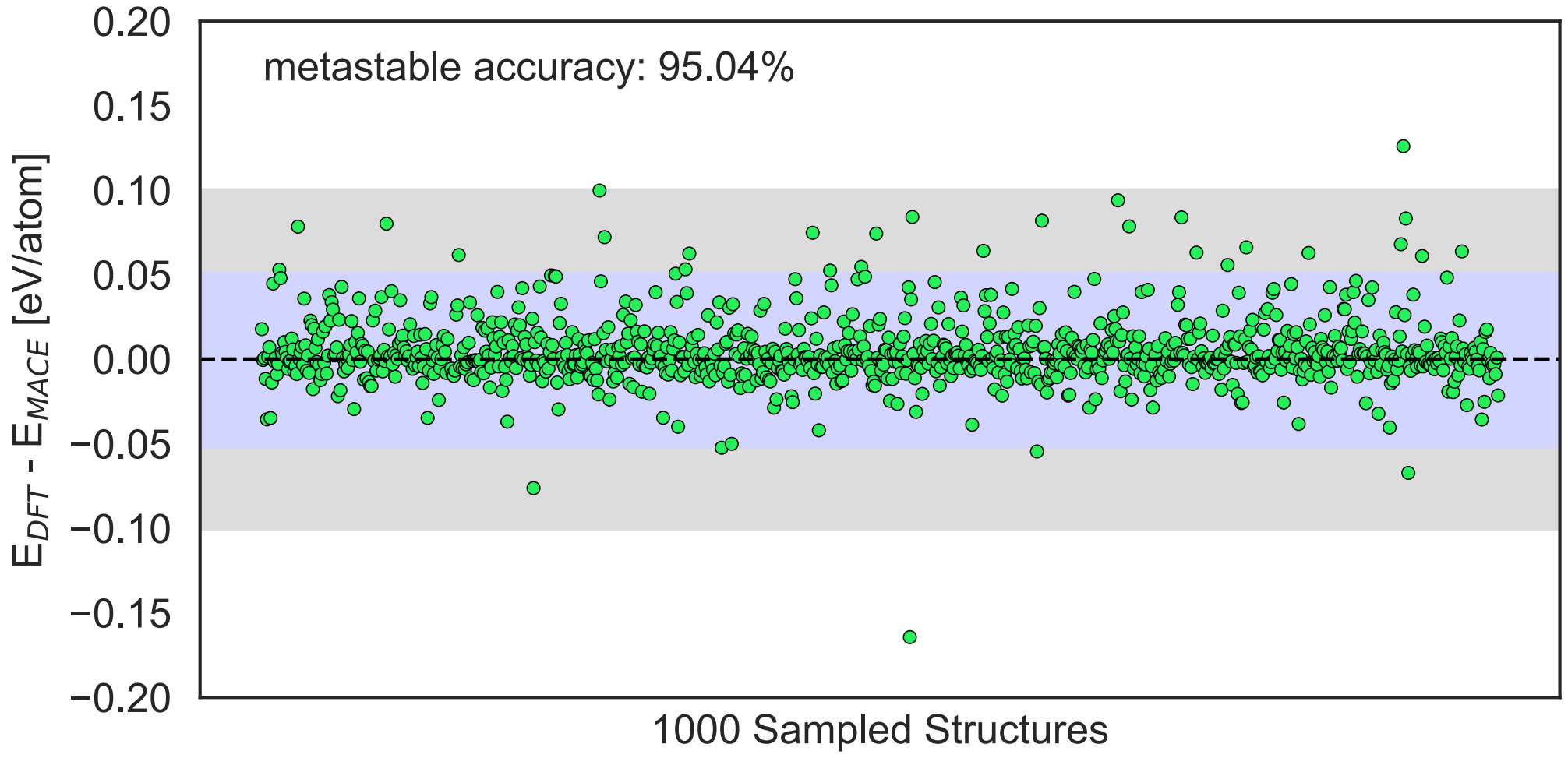

**Supplementary Fig. S1. Validation of machine learning force field accuracy for thermodynamic stability assessment.** Comparison of energy above convex hull ($E_{hull}$) values computed using MACE-MPA-0 machine learning force field ($E_{MACE}$) versus density functional theory calculations ($E_{dft}$) for 1,000 randomly generated structures. The close agreement between MLFF and DFT predictions, with errors consistently below 0.1 eV/atom within the metastability window (shaded region).

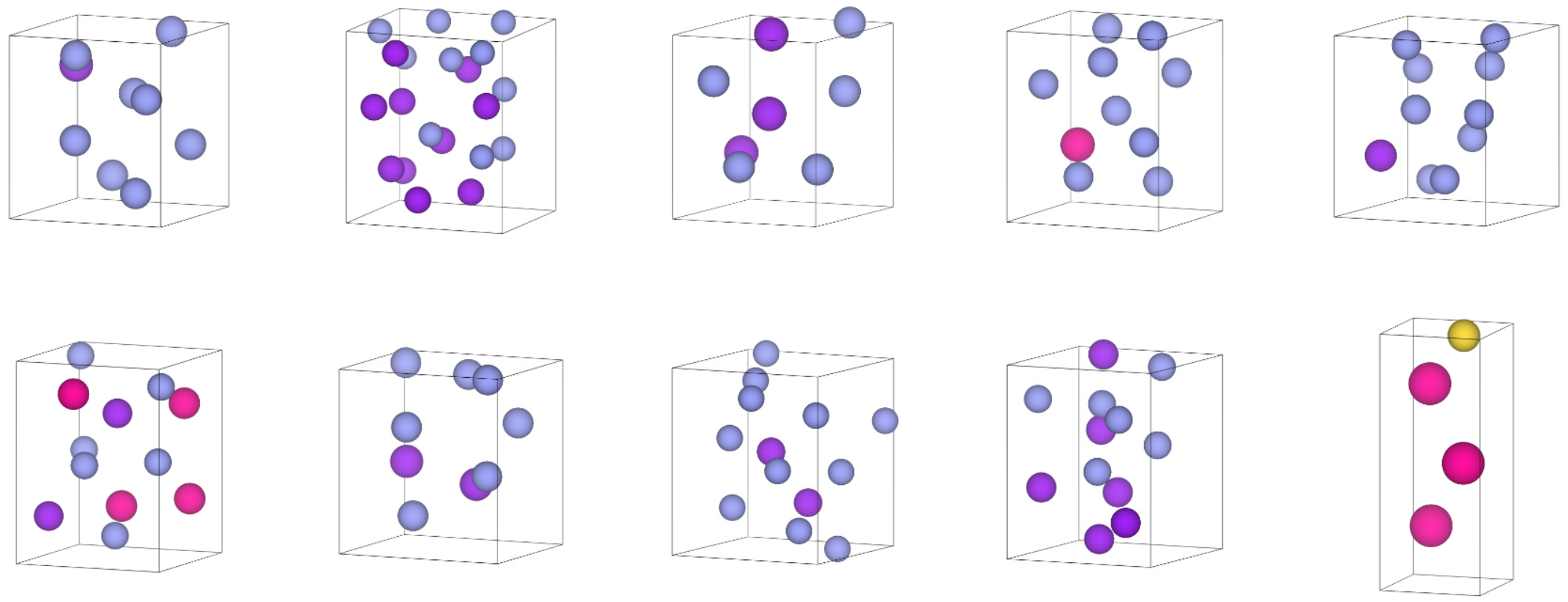

**Supplementary Fig. S2. Mode collapse during reinforcement learning without diversity rewards.** Examples of crystal structures generated by the RL-optimised model trained with creativity and stability rewards only (no diversity component).

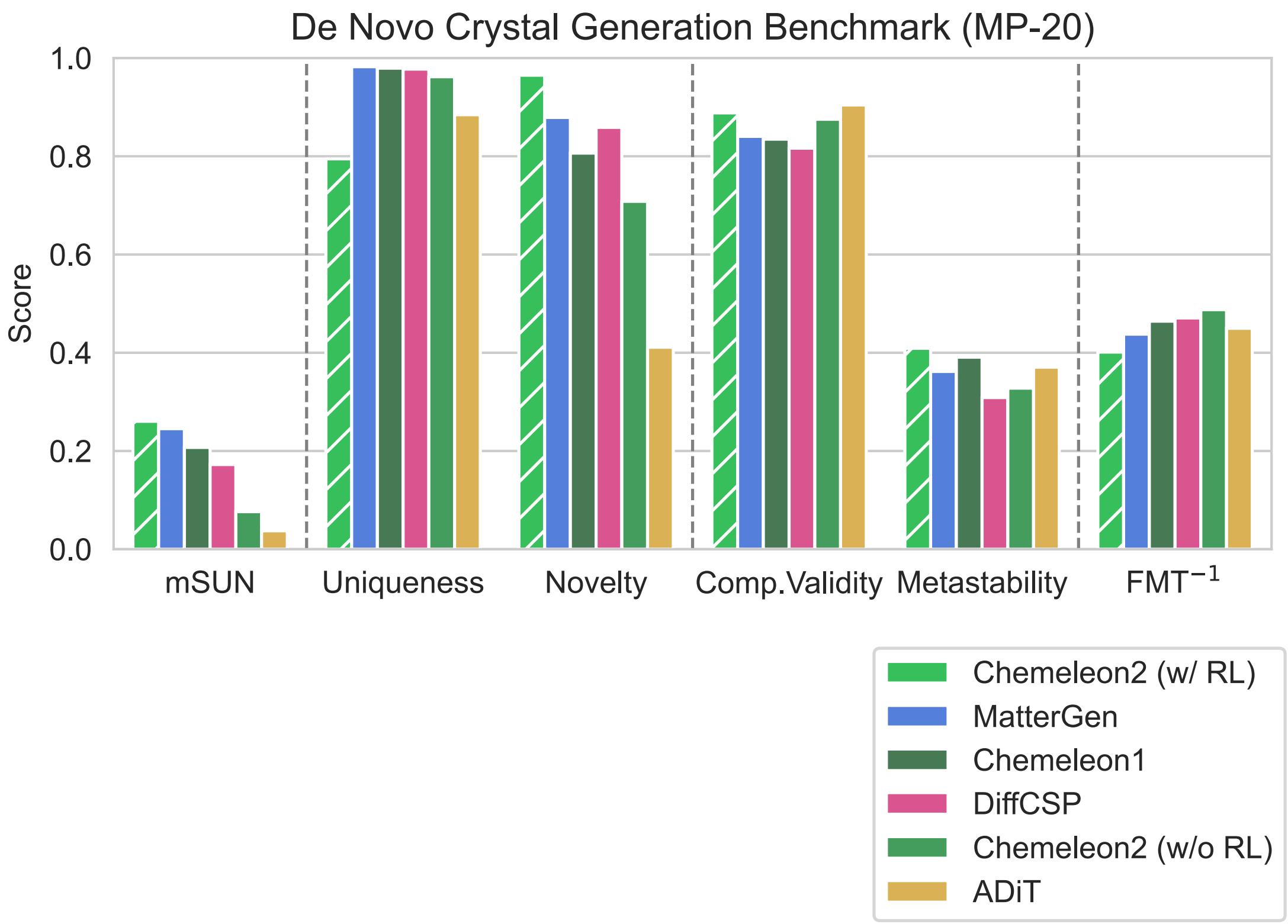

**Supplementary Fig. S3. Benchmark evaluation of generative models on the MP-20 dataset.** Comprehensive performance comparison of baseline and RL-optimised models for de novo crystal generation using 10,000 sampled structures from the standard MP-20 dataset. Metrics include mSUN (metastable, unique, and novel fraction), uniqueness, novelty, compositional validity (SMACT), metastability ($E_{hull}$ < 0.1 eV/atom), and coverage (inverse Fréchet Materials Distance, FMD$^{-1}$).

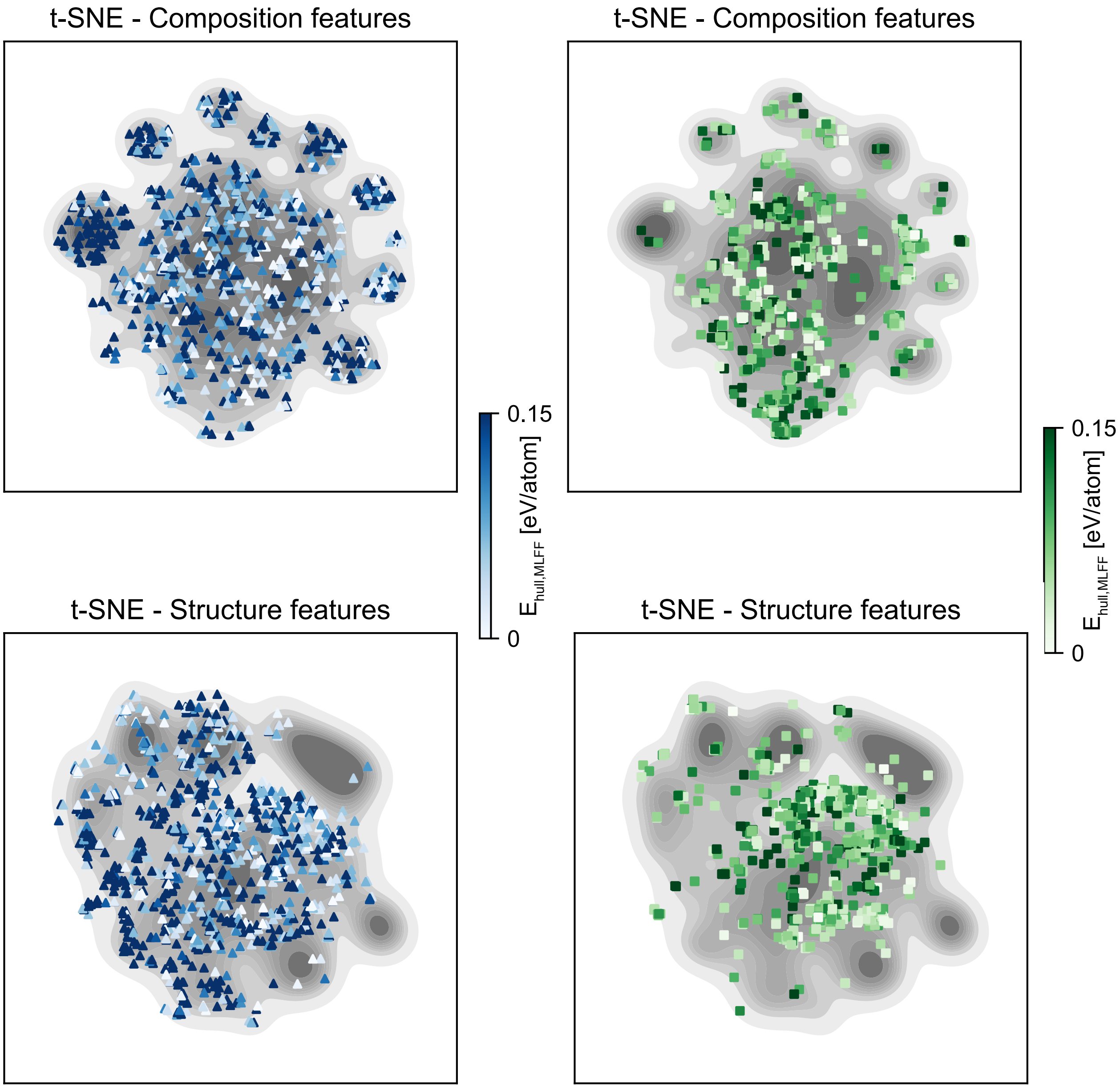

**Supplementary Fig. S4. t-SNE visualisation of generative model exploration in terms of convex hull analysis.** t-SNE projections in the first two principal components for both compositional (top row) and structural (bottom row) embedding spaces. The pre-trained Chemeleon2 baseline (left) and RL-optimised model (right) are compared. The points are coloured by energy above convex hull based on machine learning forcefields ($E_{hull,MLFF}$). Grey contours represent the density field of the Alex-MP-20 reference dataset.

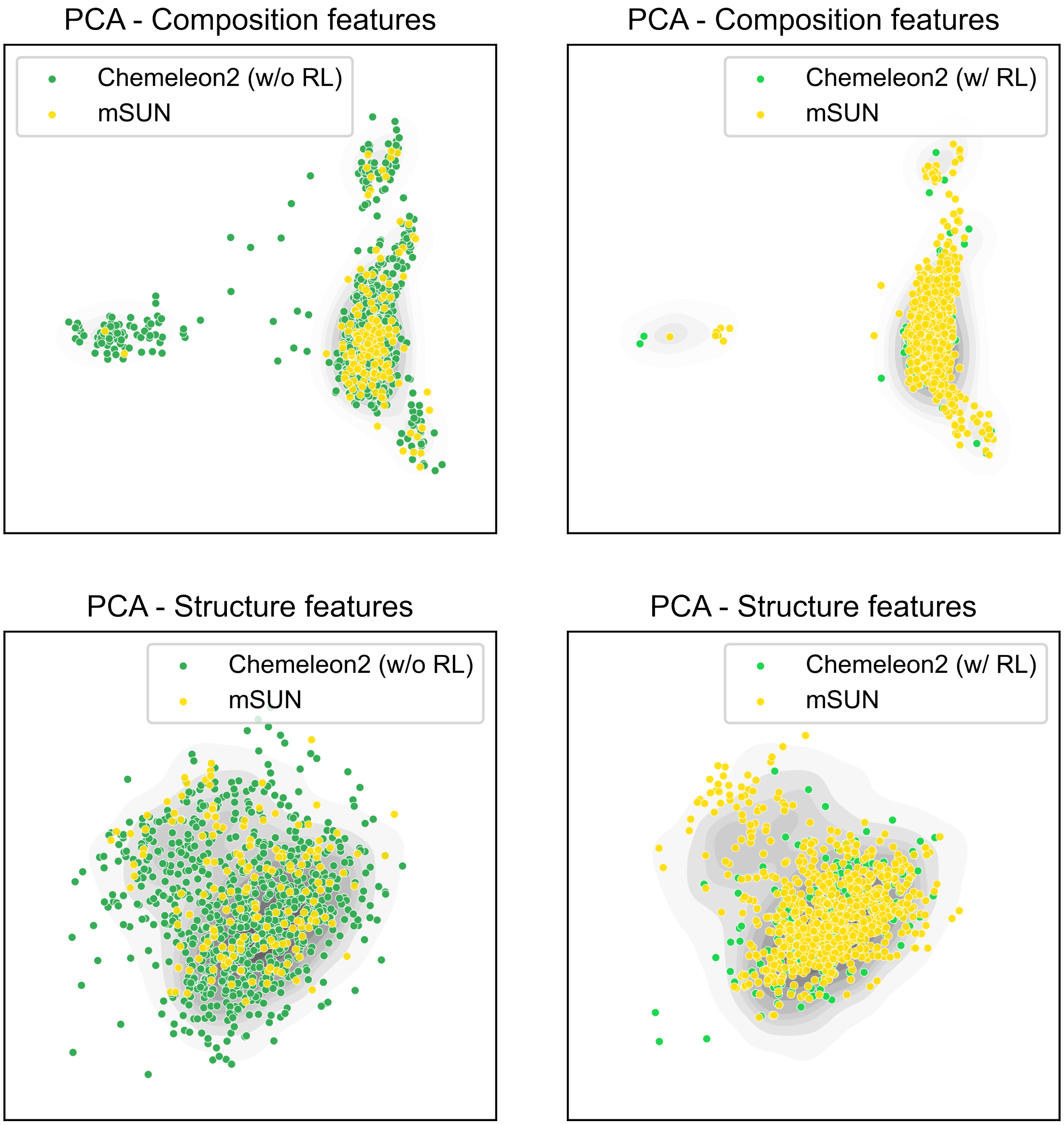

**Supplementary Fig. S5. PCA visualisation of generative model exploration.** Principal component analysis (PCA) projections in the first two principal components for both compositional (top row) and structural (bottom row) embedding spaces. The pre-trained Chemeleon2 baseline (left) and RL-optimised model (right) are compared. Grey contours represent the density field of the Alex-MP-20 reference dataset, and yellow points indicate structures satisfying the mSUN criterion.

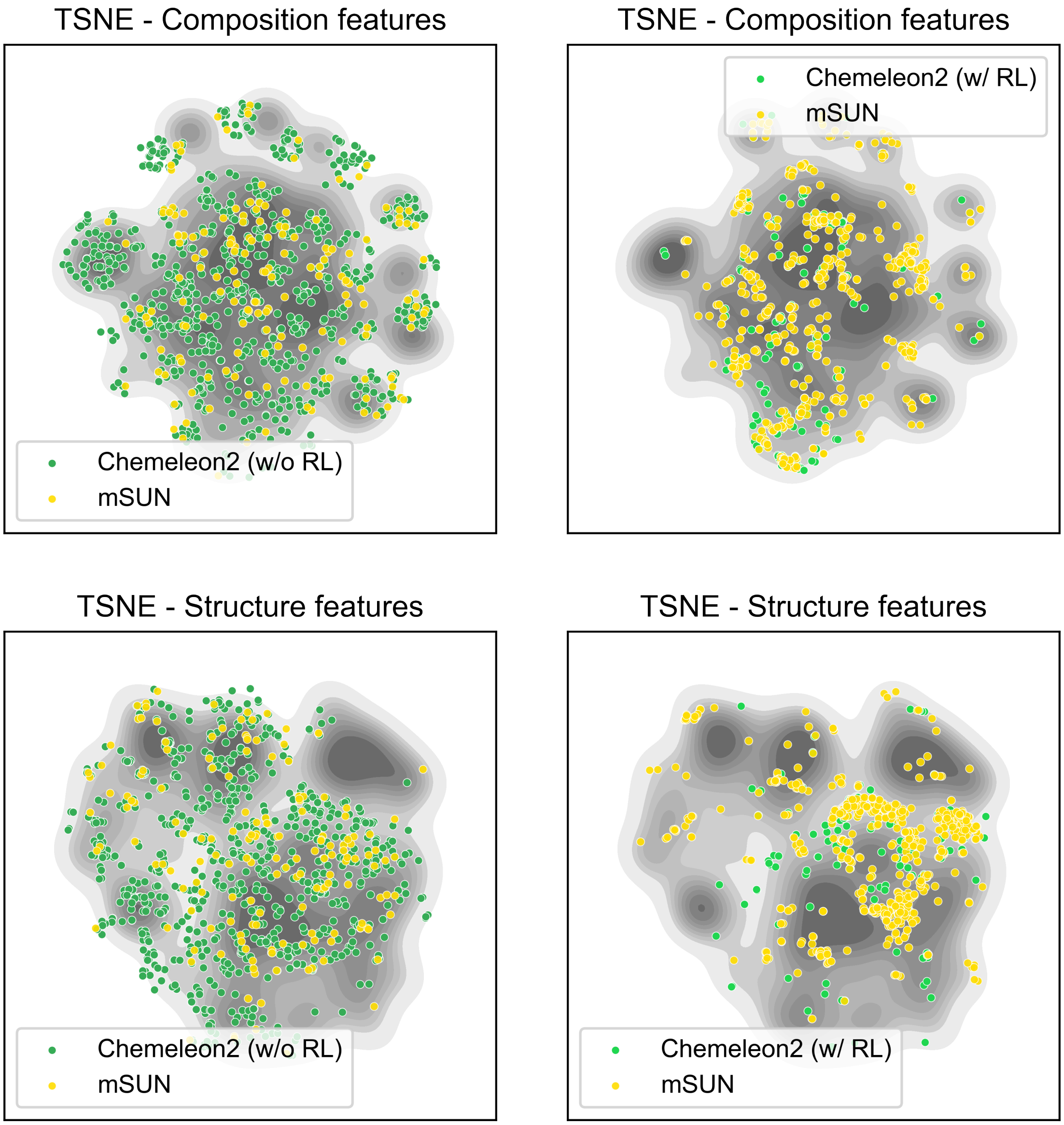

**Supplementary Fig. S6. t-SNE visualisation of generative model exploration.** t-SNE projections in the first two principal components for both compositional (top row) and structural (bottom row) embedding spaces. The pre-trained Chemeleon2 baseline (left) and RL-optimised model (right) are compared. Grey contours represent the density field of the Alex-MP-20 reference dataset, and yellow points indicate structures satisfying the mSUN criterion.

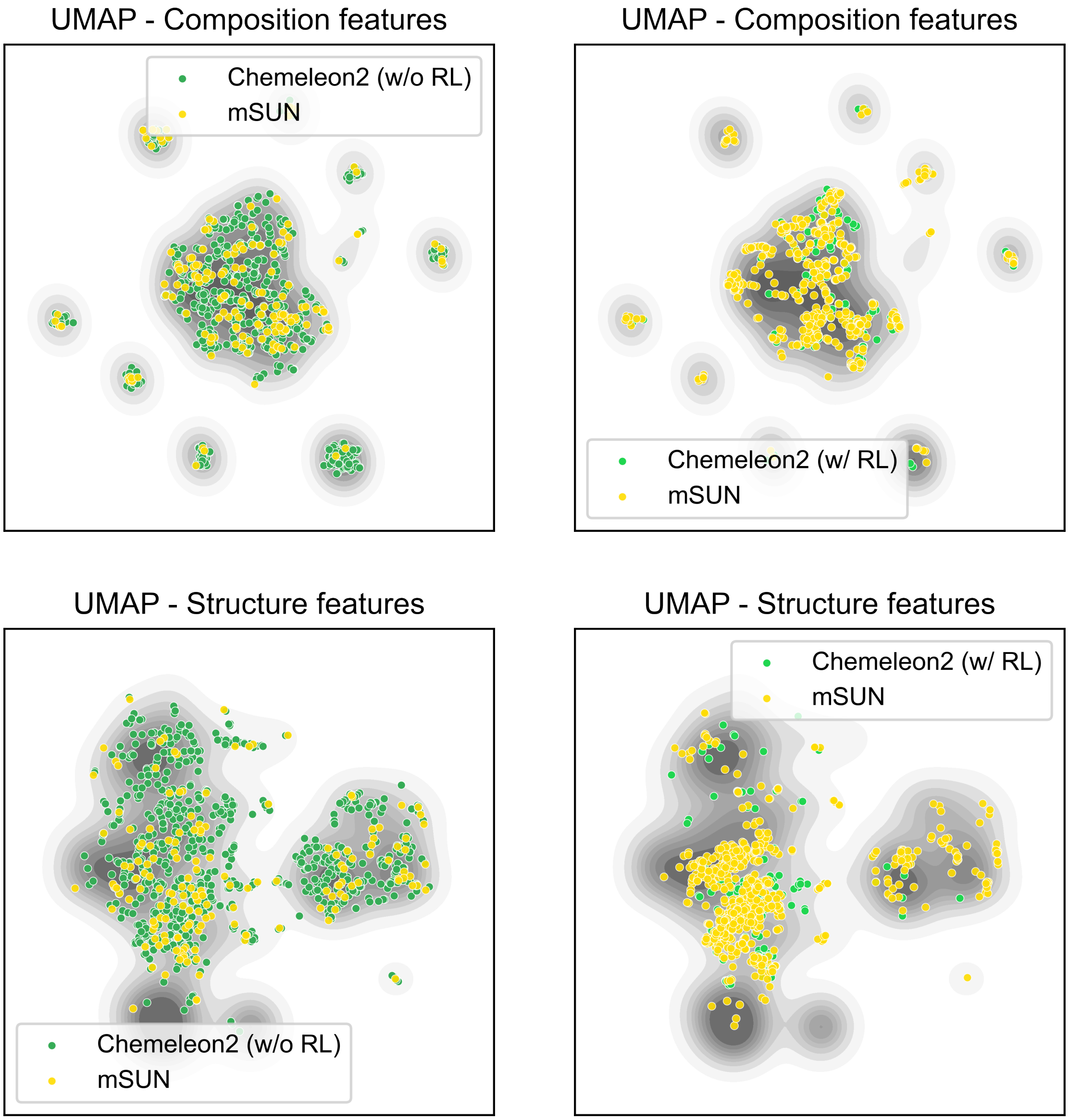

**Supplementary Fig. S7. UMAP visualisation of generative model exploration.** UMAP projections in the first two principal components for both compositional (top row) and structural (bottom row) embedding spaces. The pre-trained Chemeleon2 baseline (left) and RL-optimised model (right) are compared. Grey contours represent the density field of the Alex-MP-20 reference dataset, and yellow points indicate structures satisfying the mSUN criterion.

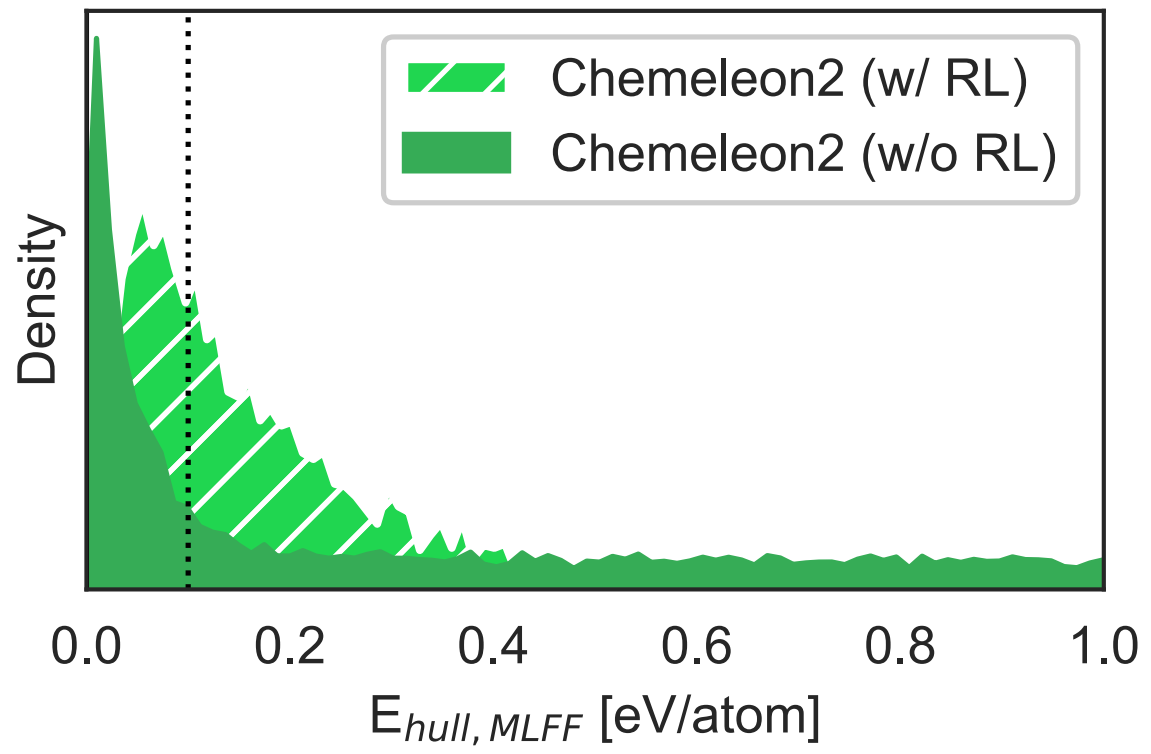

**Supplementary Fig. S8. Energy distribution analysis of generative models.** Histogram distributions of predicted energy above convex hull ($E_{hull,MLFF}$) using MACE-MPA-0 for 10,000 structures generated by baseline Chemeleon2 model and RL-optimised model trained on Alex-MP-20.

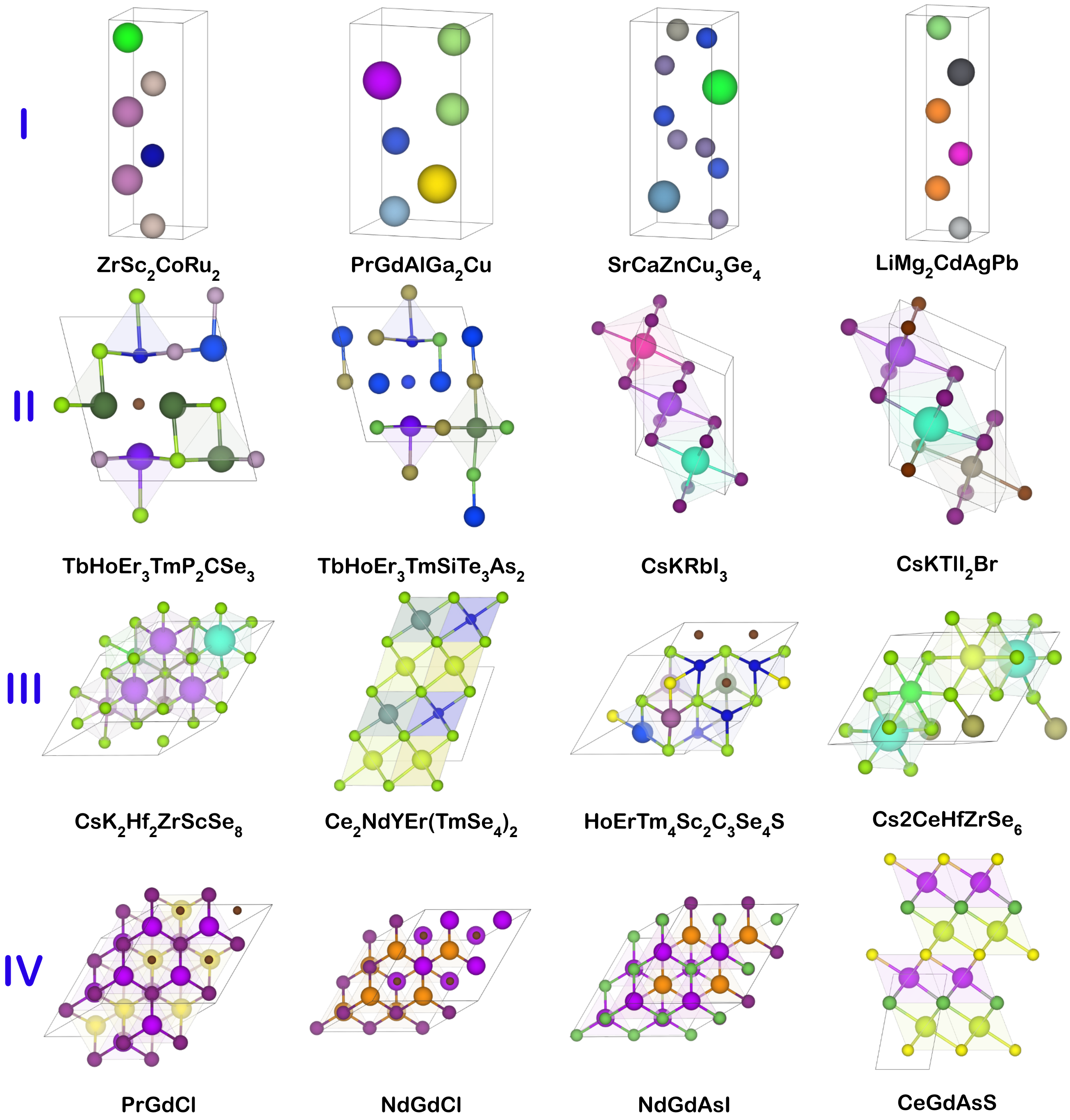

**Supplementary Fig. S9. Extended gallery of crystal structures from RL-explored chemical space.** Additional representative structures from the four primary regions identified in Fig. 4b.

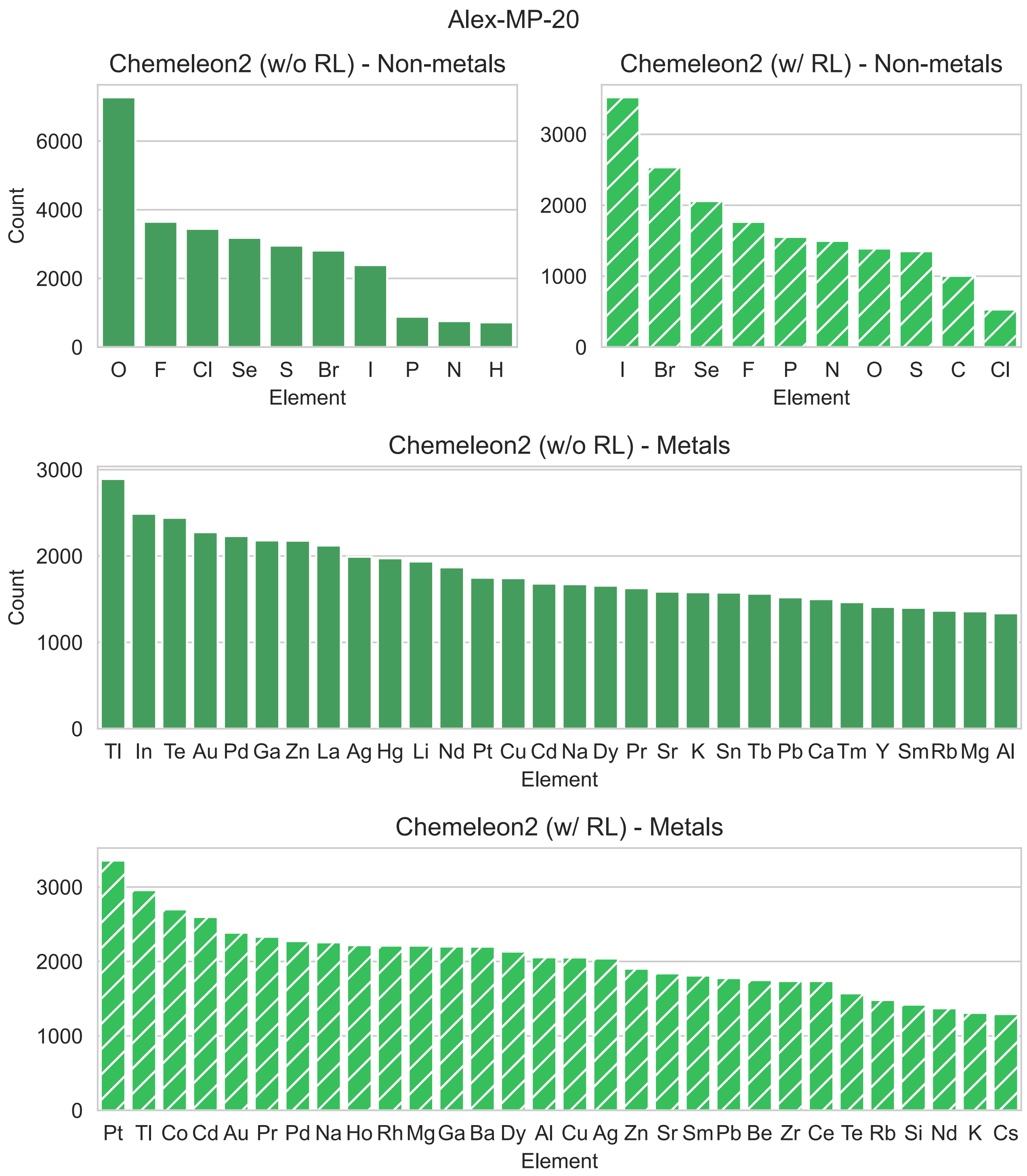

**Supplementary Fig. S10. Elemental frequency analysis for Alex-MP-20 dataset.** Ranked distribution of the most frequently observed metal and non-metal elements across 10,000 structures generated by the pre-trained Chemeleon2 baseline and RL-optimised model.

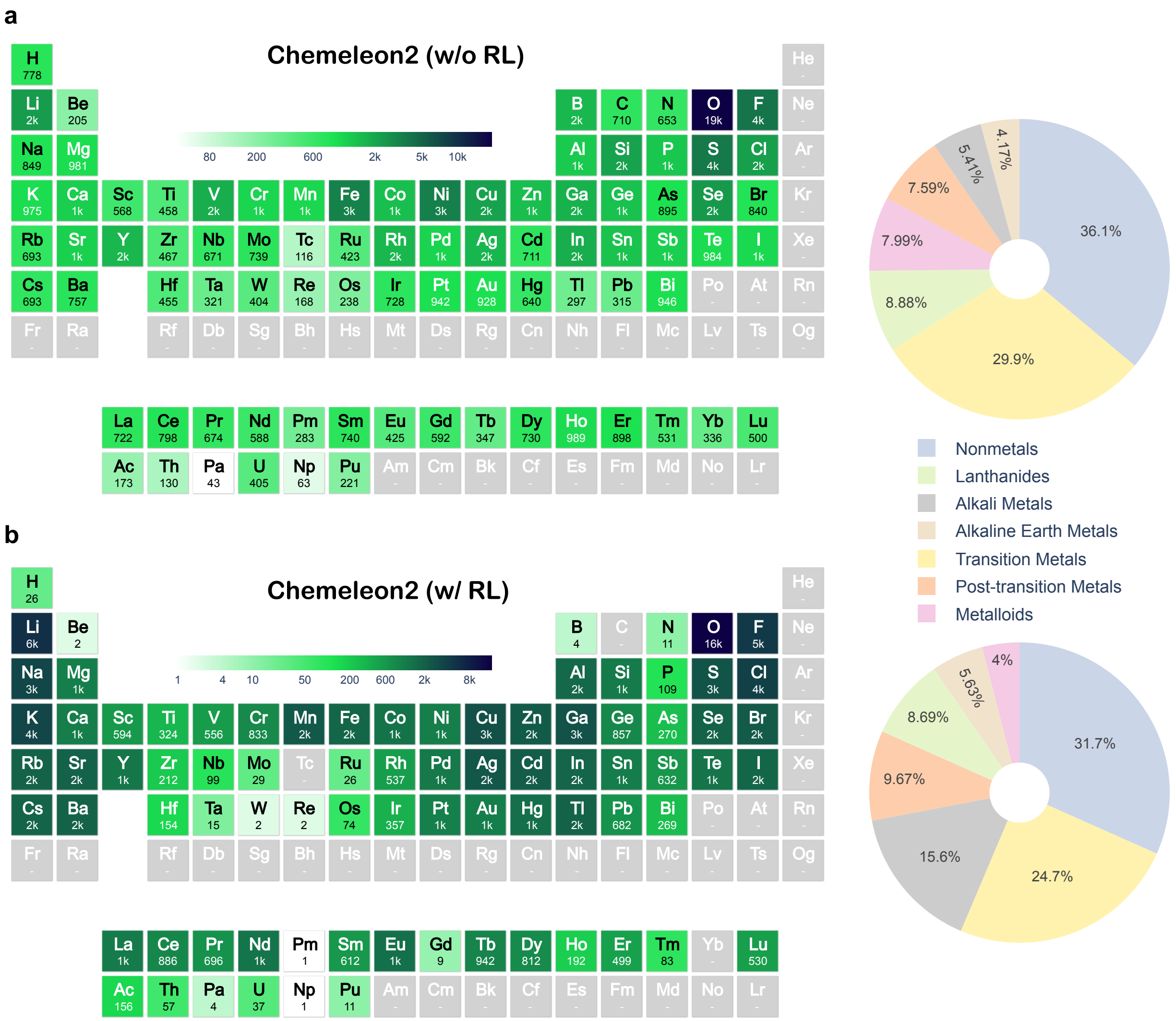

**Supplementary Fig. S11. Statistical analysis of elemental composition space for MP-20 dataset.** Elemental distribution heatmaps in periodic tables and corresponding pie charts showing the compositional statistics of 10,000 structures generated by (a) the pre-trained Chemeleon2 baseline and (b) the RL-optimised Chemeleon2 model trained on the MP-20 dataset. Color intensity represents elemental occurrence frequency, with pie charts quantifying the relative proportions across major element categories.

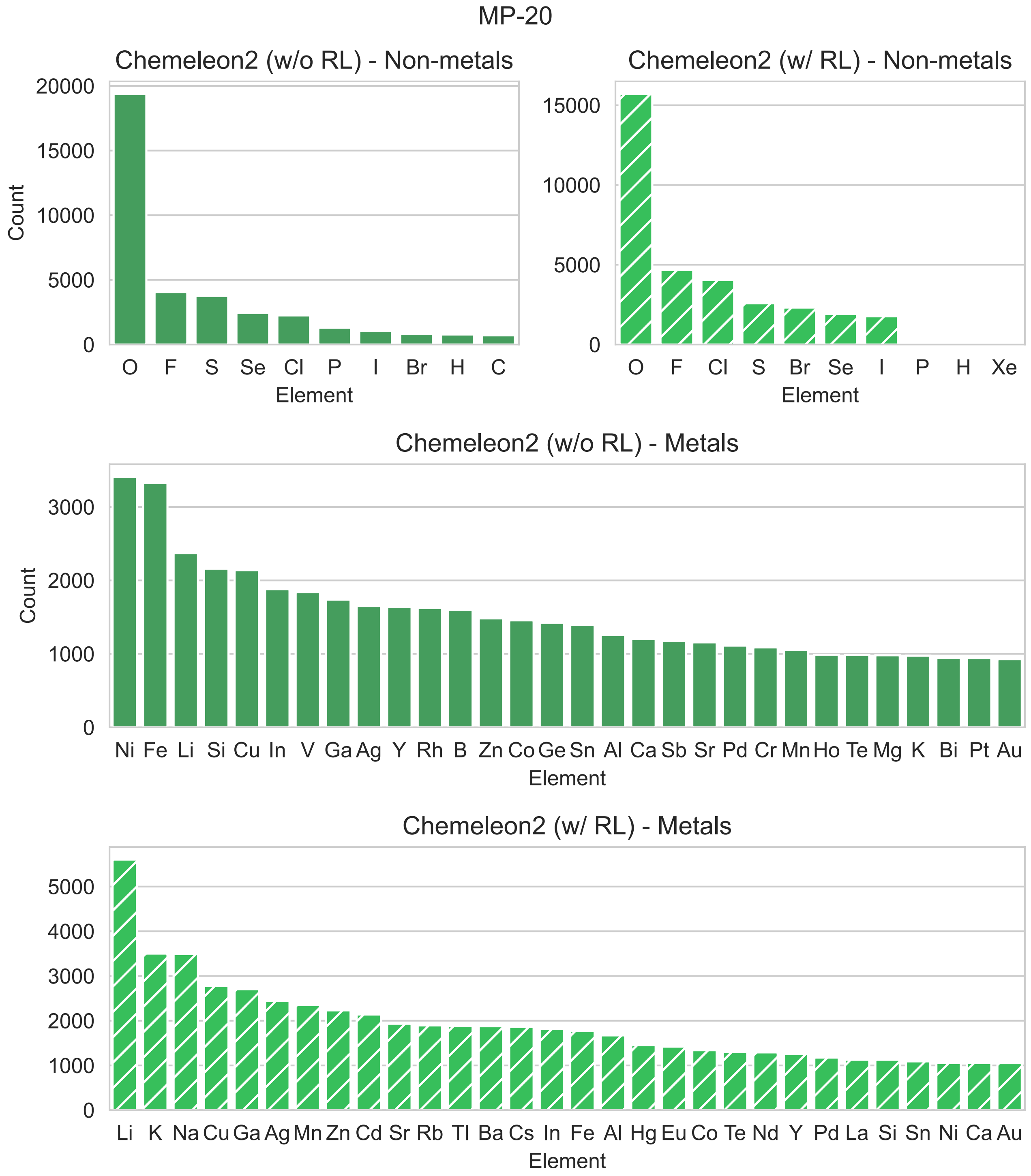

**Supplementary Fig. S12. Elemental frequency analysis for MP-20 dataset.** Ranked distribution of the most frequently observed metal and non-metal elements across 10,000 structures generated by the pre-trained Chemeleon2 baseline and RL-optimised model.

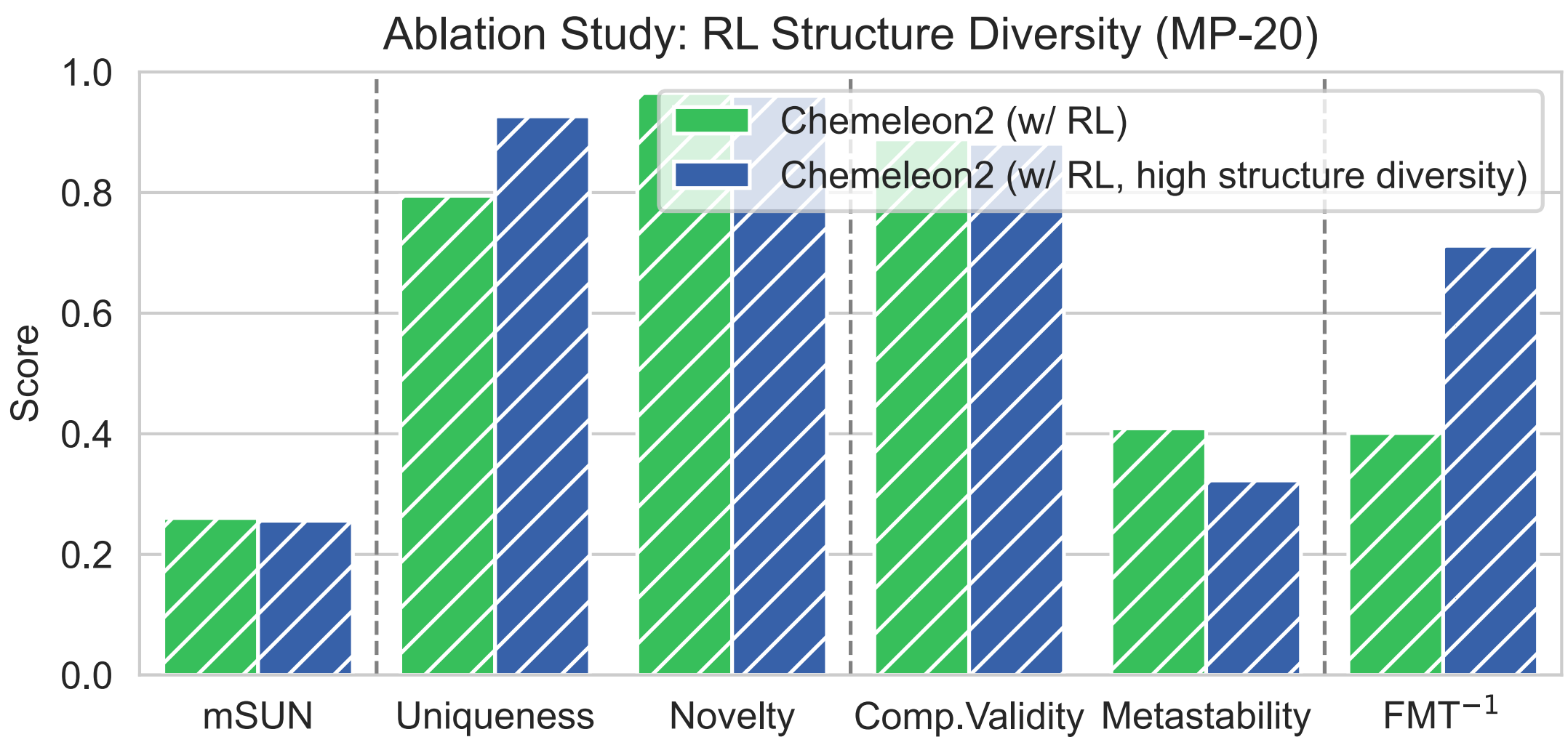

**Supplementary Fig. S13. Ablation study of structural diversity reward.** Benchmark comparison of de novo crystal generation performance for Chemeleon2 models fine-tuned via reinforcement learning with two distinct diversity reward configurations: conservative weighting (0.1) versus elevated weighting (1.0) for structural diversity.

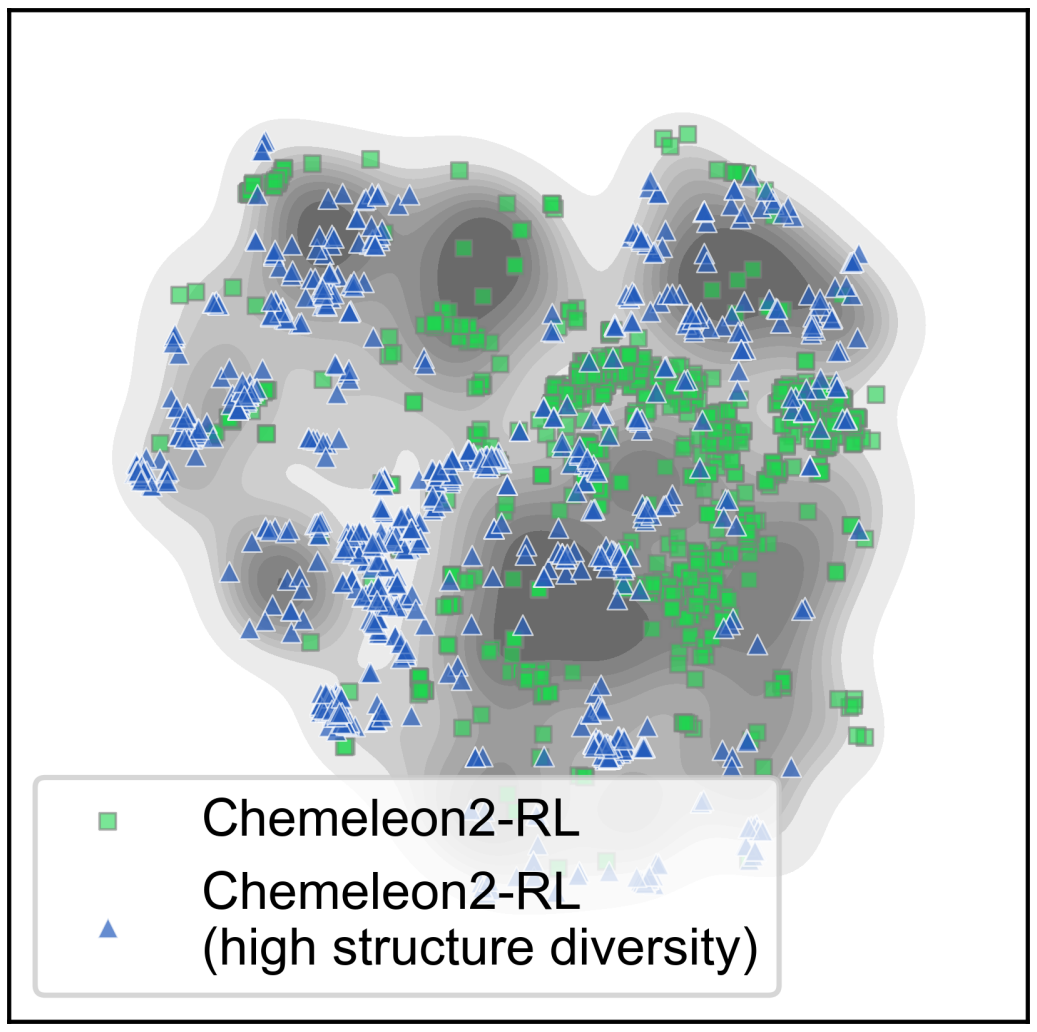

**Supplementary Fig. S14. t-SNE visualisation comparing structural diversity reward weighting strategies.** t-SNE projections of structural embeddings for 1,000 generated structures from Chemeleon2 models optimised with conservative diversity weighting (0.1) and (b) elevated diversity weighting (1.0) for structural diversity, coloured by green and blue, respectively. Grey contours represent the reference MP-20 distribution density.

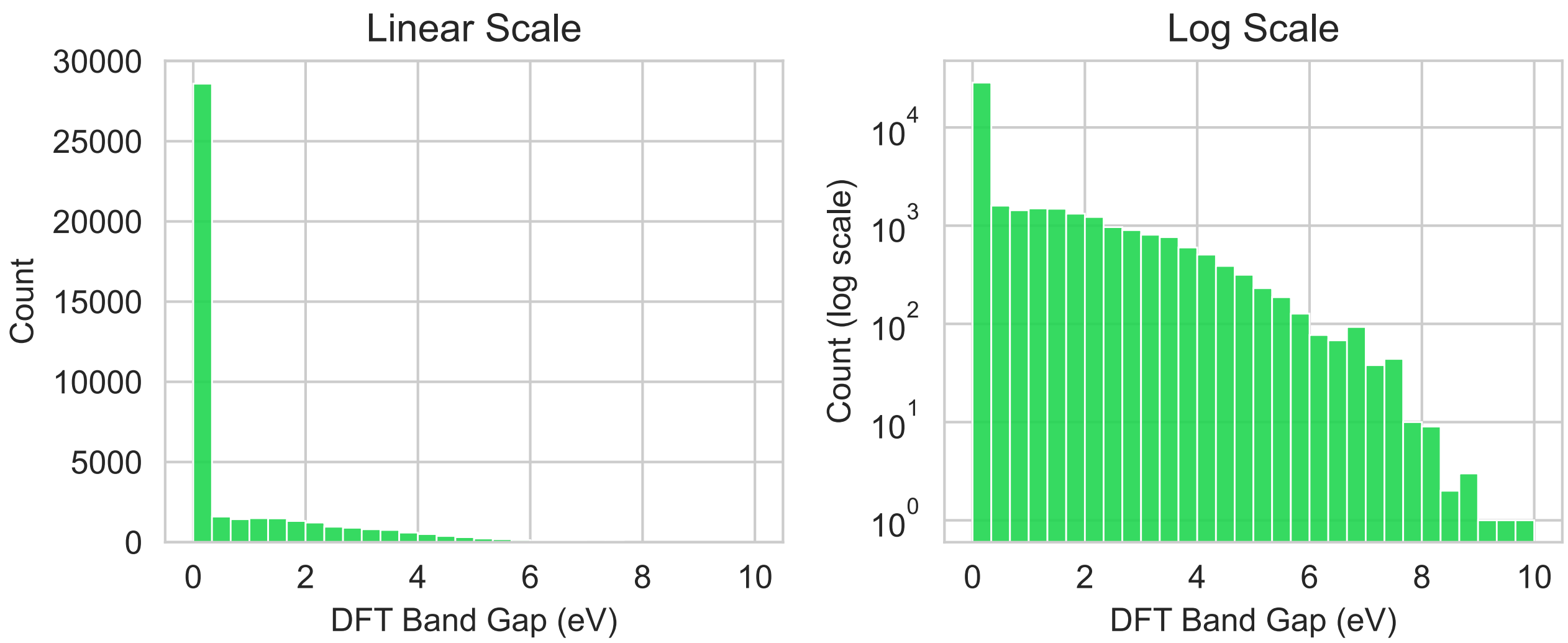

**Supplementary Fig. S15. Target bandgap distribution in Alex-MP-20 dataset.** Histogram showing the distribution of DFT-calculated electronic bandgaps across 43,295 structures in the Alex-MP-20 training dataset, in terms of linear scale (left) and log scale (right).

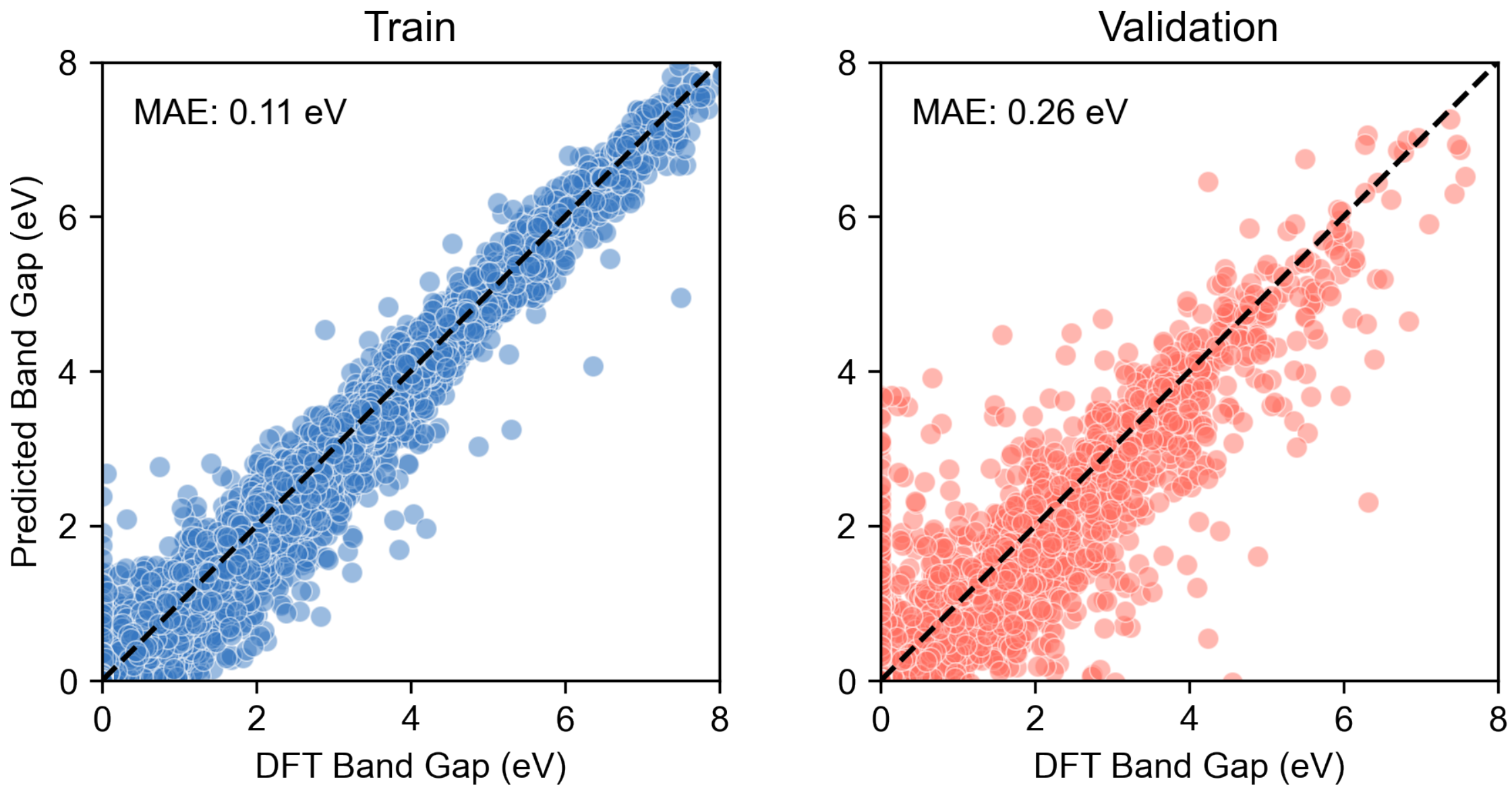

**Supplementary Fig. S16. Bandgap surrogate model performance validation.** Predicted from the bandgap surrogate model using VAE versus true bandgap values for the (a) training set and (b) validation set.

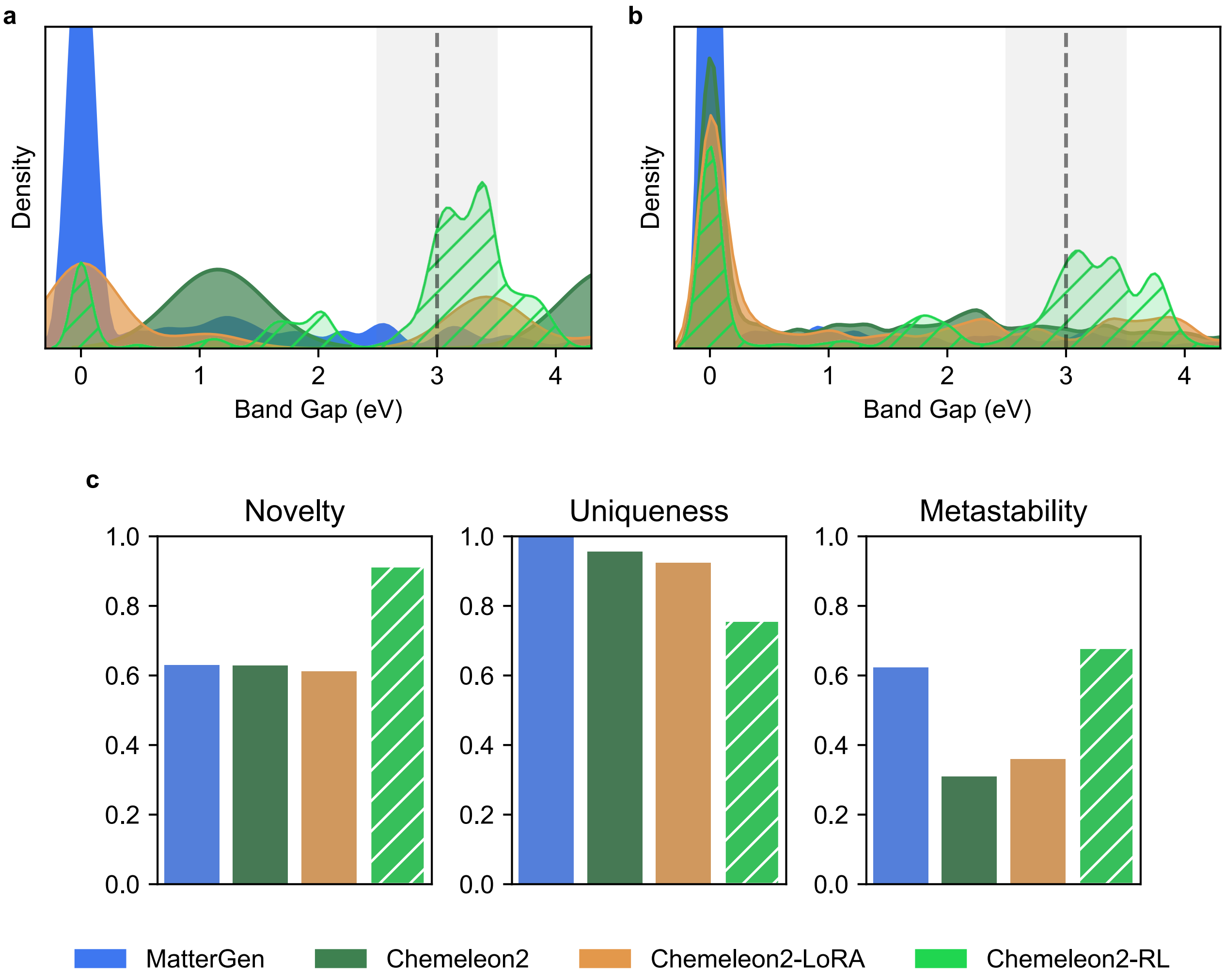

**Supplementary Fig. S17. Property-guided generation benchmark.** (a) DFT-calculated Bandgap distributions for structures satisfying the mSUN criterion, showing reduced zero-gap artifacts compared to unfiltered results. (b) Bandgap distributions for 512 sampled structures without any filter. (c) Component analysis of mSUN score across CFG, LoRA-adapted CFG, and RL-based approaches for both Chemeleon2 and MatterGen.

**Supplementary Table S1. Quantitative benchmark results.** Performance of baseline and RL-optimised generative models in de novo generation of crystal structures for 10,000 samples trained with Alex-MP-20 and MP-20 dataset.

| dataset | Model | mSUN | Uniqueness | Novelty | Comp. Validity | Meta-stability | $FMT^{-1}$ |
|---|---|---|---|---|---|---|---|
| MP-20 | DiffCSP | 17.2 | 97.7 | 85.9 | 81.6 | 30.9 | 47.1 |
| | ADiT | 3.8 | 88.5 | 41.1 | 90.4 | 37.1 | 45.0 |
| | MatterGen | <u>24.5</u> | 98.2 | 87.9 | 84.0 | 36.2 | 43.8 |
| | Chemeleon1 | 20.7 | 97.9 | 80.6 | 83.5 | 39.1 | 46.4 |
| | Chemeleon2 | 7.6 | 96.2 | 70.8 | 87.5 | 32.8 | 48.8 |
| | Chemeleon2 (w/ RL) | **26.0** | 79.5 | 96.5 | 88.8 | 40.9 | 40.1 |
| Alex-MP-20 | MatterGen | <u>41.0</u> | 99.8 | 68.8 | 87.0 | 71.3 | 33.0 |
| | Chemeleon1 | 32.3 | 99.6 | 70.8 | 86.7 | 60.8 | 33.5 |
| | Chemeleon2 | 15.9 | 99.4 | 62.3 | 89.6 | 51.2 | 32.3 |
| | Chemeleon2 (w/ RL) | **61.3** | 88.7 | 97.5 | 94.8 | 72.1 | 52.8 |